\documentclass[journal]{IEEEtran}
\ifCLASSINFOpdf
\else
\fi
\usepackage[pdftex]{graphicx}
\usepackage{subfigure}
\usepackage{amsmath}
\usepackage{stfloats}
\usepackage{mathtools}
\usepackage{cite}
\usepackage{subeqnarray}
\usepackage{cases}
\usepackage{mathrsfs}
\usepackage{graphicx}
\usepackage{threeparttable}
\usepackage{acronym}
\usepackage{caption}
\usepackage{dsfont}
\usepackage{url}

\newtheorem{problem}{Problem}

\begin{document}
%
% paper title
% Titles are generally capitalized except for words such as a, an, and, as,
% at, but, by, for, in, nor, of, on, or, the, to and up, which are usually
% not capitalized unless they are the first or last word of the title.
% Linebreaks \\ can be used within to get better formatting as desired.
% Do not put math or special symbols in the title.
\title{Stop Line Aided Cooperative Positioning \\ of Connected Vehicles}
%
%
% author names and IEEE memberships
% note positions of commas and nonbreaking spaces ( ~ ) LaTeX will not break
% a structure at a ~ so this keeps an author's name from being broken across
% two lines.
% use \thanks{} to gain access to the first footnote area
% a separate \thanks must be used for each paragraph as LaTeX2e's \thanks
% was not built to handle multiple paragraphs
%

\author{Xingqi~Wang,
        Chaoyang~Jiang,
        Shuxuan~Sheng,
        Yanjie~Xu,
        and~Yifei~Jia% <-this % stops a space
\thanks{This work was supported by the National Natural Science Foundation of China(No.52002026, No.U1764257, and No.U20A20333)(\textit{Corresponding author: Chaoyang Jiang}).

The authors are with the School of Mechanical Engineering, Beijing Institute of Technology, Beijing, China, 100081 (e-mail:wxq1115@foxmail.com; cjiang@bit.edu.cn;)}% <-this % stops a space
% \thanks{J. Doe and J. Doe are with Anonymous University.}% <-this % stops a space
%\thanks{Manuscript received April 19, 2005; revised August 26, 2015.}
}

\maketitle

% As a general rule, do not put math, special symbols or citations
% in the abstract or keywords.
\begin{abstract}
This paper develops a stop line aided cooperative positioning framework for connected vehicles, which creatively utilizes the location of the stop-line to achieve the positioning enhancement for a vehicular ad-hoc network (VANET) in intersection scenarios via Vehicle-to-Vehicle (V2V) communication. Firstly, a self-positioning correction scheme for the first stopped vehicle is presented, which applied the stop line information as benchmarks to correct the GNSS/INS positioning results. Then, the local observations of each vehicle are fused with the position estimates of other vehicles and the inter-vehicle distance measurements by using an extended Kalman filter (EKF). In this way, the benefits of the first stopped vehicle are extended to the whole VANET. Such a cooperative inertial navigation (CIN) framework can greatly improve the positioning performance of the VANET. Finally, experiments in Beijing show the effectiveness of the proposed stop line aided cooperative positioning framework.  
\end{abstract}

% Note that keywords are not normally used for peerreview papers.

% For peer review papers, you can put extra information on the cover
% page as needed:
% \ifCLASSOPTIONpeerreview
% \begin{center} \bfseries EDICS Category: 3-BBND \end{center}
% \fi
%
% For peerreview papers, this IEEEtran command inserts a page break and
% creates the second title. It will be ignored for other modes.
\IEEEpeerreviewmaketitle

\section{Introduction}
% The very first letter is a 2 line initial drop letter followed
% by the rest of the first word in caps.
% 
% form to use if the first word consists of a single letter:
% \IEEEPARstart{A}{demo} file is ....
% 
% form to use if you need the single drop letter followed by
% normal text (unknown if ever used by the IEEE):
% \IEEEPARstart{A}{}demo file is ....
% 
% Some journals put the first two words in caps:
% \IEEEPARstart{T}{his demo} file is ....
% 
% Here we have the typical use of a "T" for an initial drop letter
% and "HIS" in caps to complete the first word.
\IEEEPARstart{R}{eliable} and accurate positioning is essential for path planning, trajectory tracking, and the safe operation of autonomous vehicles in all traffic conditions \cite{Reid19}. Intersection scenarios are of great significance for autonomous driving.  The maturing vehicle-to-everything (V2X) communication has the potential to upgrade separate autonomous vehicles to connected vehicles, which constitute vehicular ad hoc networks and can greatly improve their positioning performance, road safety and passenger convenience, especially for intersection scenarios \cite{Far20}. Hence, positioning for connected vehicles in intersection scenarios has increasingly attracted recent attention.  

 Global navigation satellite system (GNSS) can provide the global position for a vehicle with a GNSS receiving terminal, which is  broadest-used for vehicle positioning. But its performance in highly urbanized areas is severely degraded due to multi-path effects \cite{Wen19, Wen21, Zhang21}, and the GNSS signal can even be denied in some special areas. With the help of the real-time kinematic (RTK) technique, GNSS can provide more accurate results. Integrated with an inertial navigation system (INS), both accuracy and availability of the positioning information can be enhanced \cite{Liu18, Nour18}. However, the current GNSS/INS integrated navigation systems with RTK still cannot meet the safety requirements of autonomous vehicles in many urbanized scenarios \cite{Reid19, Cao20, Li19-IF,Jiang19-tvt}. Fortunately, rich environmental information is available in urbanized areas such as intersection areas, and 5G communication infrastructure is being rapidly built in many countries like China. Both the environmental correction and the cooperative positioning based on the V2X communication have the potential to improve the accuracy and reliability of vehicular positioning, especially in GNSS-challenged areas \cite{Suhr17, Zhu20, Li21-ITS}.

Environmental information such as lane lines, traffic signs, and pedestrians can be exploited to enhance vehicle position estimation \cite{Suhr17, Jiang13, Xu20}.  Suhr et al. \cite{Suhr17} introduced the observations of symbolic road markings into a GNSS/INS integrated navigation system via a particle filter with which the localization error was decreased. Lanes were formulated to constraints and introduced to a constrained Kalman filter in \cite{Jiang13, Xu20}, which significantly  reduced the lateral errors. Wang et al. \cite{Wang19-ITS}  fused the visually recognized traffic lights with the INS information and enhanced the vehicle localization in an intersection scenario. Welzel et al. \cite{Welzel15} utilized the location of traffic signs to correct the GNSS measurements in a Bayesian filtering framework. Similarly, Qu et al. \cite{Qu15}  applied geo-referenced traffic signs and developed a localization enhancement method that reduced the accumulated drifts of visual sensors. Considering the environmental information, all the works mentioned above can provide remarkable improvement in terms of positioning accuracy. However, they all focused on self-positioning and has no help for the vehicles that cannot observe the particular environmental markings.

With the V2X communication, vehicles can interact with other vehicles and receive information from roadside units (RSU), which allows cooperative positioning and improves the overall positioning performance of the VANET \cite{Alam13, Xiong21}. 
Song et al. \cite{Song20} applied the traffic signs as benchmarks to correct the GNSS observations with the aid of laser radar, and then broadcast the estimated GNSS error to all vehicles in the VANET.  Alam et al. \cite{Alam12} and Li et al. \cite{Li19} developed an RSU-assisted GNSS positioning method to achieve lane-level positioning. They identified the driving lane by utilizing the carrier frequency offset (CFO) and received signal strength (RSS) broadcast from roadside infrastructure beacons, respectively. Chen et al. \cite{Chen19} utilized the observations of roadside cameras via vehicle-to-infrastructure (V2I) communication to correct the GNSS positioning, and transmitted the GNSS correction information to the connected vehicles via V2V communications. However, these studies do not consider any inter-vehicle measurements that are significant for the error reduction of cooperative positioning. In \cite{Fas18}, the azimuths of partial vehicles with a single active RSU were observed. Both V2V and V2I communications were applied in a novel EKF-based cooperative positioning algorithm and then the positioning performance of all vehicles in the VANET is improved.

However,the RSUs introduce additional costs, which limits their density in urban areas and the practical application shortly. Hence, it is interesting to consider the cooperative positioning for connected vehicles in the RSUs-absent environment. Hoang et al. \cite{Hoang19} built a V2V communication network via ultra-wideband (UWB) devices and integrated the local GNSS observations with the vehicle-to-vehicle distance measurements, which provided a remarkable positioning enhancement. Meyer et al. \cite{Meyer16} and Soatti et al. \cite{Soatti18} presented an implicit cooperative positioning algorithm and fused the local features such as pedestrians, traffic lights, and inactive vehicles by a consensus methodology to refine vehicular positioning of a VANET. Liu et al. \cite{Liu17} integrated the dedicated short-range communication (DSRC) and GNSS observations and applied a modified robust cubature Kalman filter to improve the cooperative positioning robustness and adaptive performance. Wang et al. \cite{Wang13} and Yang et al. \cite{Yang20} presented a two-layer structure for a multi-sensor multi-vehicle localization scenario, in which the local filter integrated local onboard sensor measurements and the global filter fused the observations from other vehicles. Simulation results demonstrated that the localization accuracy could be greatly improved by cooperation among multiple sensors and vehicles.

In this work, we apply the location of known stop-lines to correct the position estimation of the first stopped vehicle and extend the enhancement to all vehicles in the VANET via a well-designed cooperative inertial navigation (CIN) framework. The onboard observations of each vehicle, the environmental information (i.e., the location of stop-lines), and the inter-vehicle distance observations are fused for positioning enhancement of the whole VANET. The main contributions of this paper are as follows:

(1)The stop lines are utilized to assist the GNSS/INS integrated positioning system of the first stopped vehicle, which not only corrects the estimation of the self-position but also provides a better initialization for the positioning filter after the traffic light turns green. To the best of our knowledge, it is the first work for vehicle positioning enhancement with the help of stop lines.

(2)We apply the V2V communications and the inter-vehicle distance measurements to extend the improvement of the first stopped vehicle to the whole VANET. The proposed stop line aided cooperative positioning framework can improve the vehicle positioning performance for intersection scenarios.

(3)UWB sensors are used to build the V2V communication networks and to measure the inter-vehicle distances. Experimental results show that the proposed framework remarkably improves the positioning performance of the vehicles in an intersection scenario, especially for the vehicles directly communicating with the first stopped vehicle. 

The rest of this paper is organized as follows. The problem formulation is given in Section II. Section III presents the stop line aided self-positioning correction scheme in detail. In Section IV, we propose the cooperative inertial navigation framework. The experimental results and analysis of the proposal are discussed in section V. Finally, Section VI concludes this paper.

\section{Problem Statement}
\begin{figure}[!t]
\centering
\includegraphics[scale=0.7]{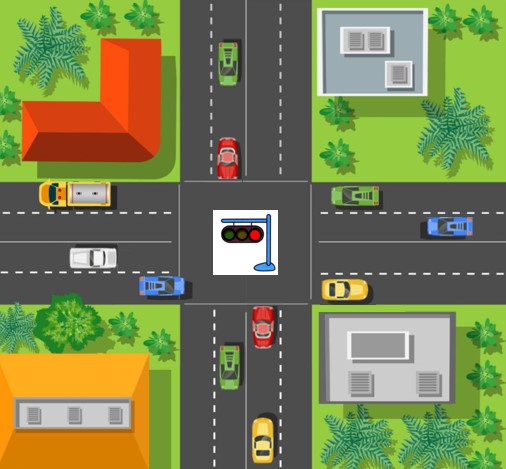}
% where an .eps filename suffix will be assumed under latex, 
% and a .pdf suffix will be assumed for pdflatex; or what has been declared
% via \DeclareGraphicsExtensions.
\caption{A typical intersection scenario for connected vehicles.}
\label{fig1}
\end{figure}

In high-traffic intersection scenarios, vehicles can be classified into moving vehicles and stopped vehicles. According to the traffic rules, the first stopped vehicle is usually next to the stop line, as the two red cars in Fig.\ref{fig1}. The locations of the stop lines are commonly \textit{a prior} known. The first stopped vehicle will be directly localized if we can obtain the relative location with the stop line. Even if the relative location is unknown and simply assumed that the first stopped vehicle is close to the stop line, we can still estimate the longitudinal position, which can be used to enhance positioning results of the GNSS/INS integrated navigation system especially in highly urbanized areas.  

It is known that with V2V communications and inter-vehicle distance and/or angle measurements, the positioning performance of the vehicles in a VANET can be enhanced. The first stopped vehicle can be viewed as an anchor, and we hope that the positioning improvement of the first stopped vehicle can be extended via a cooperative positioning algorithm. Therefore, in this paper, we focus on the following two problems.

\begin{problem}
For intersection scenarios, given the observations of GNSS and INS of the first stopped vehicle, the location of the stop lines, how can we estimate the position of the first stopped vehicle from the GNSS/INS integrated navigation system, and enhance the estimation with the aid of the stop line? 
\end{problem}

\begin{problem}
In intersection scenarios, the enhanced position estimation of the first stopped vehicle, the V2V communication, and the distance measurements of neighbor inter-vehicles are available. With the aid of the positioning enhancement of the first stopped vehicle, design a cooperative positioning algorithm to improve the position estimation of all vehicles in the VANET. 
\end{problem}

\section{Stop line aided self-positioning enhancement}

In this section, we first introduce a GNSS/INS integration method for vehicle self-positioning and then develop a stop line aided GNSS/INS integration solution to enhance the positioning performance of the first stopped vehicle in intersection scenarios.

\subsection{GNSS/INS integration for vehicle positioning}

\begin{figure}[!t]
\centering
\includegraphics[scale=0.3]{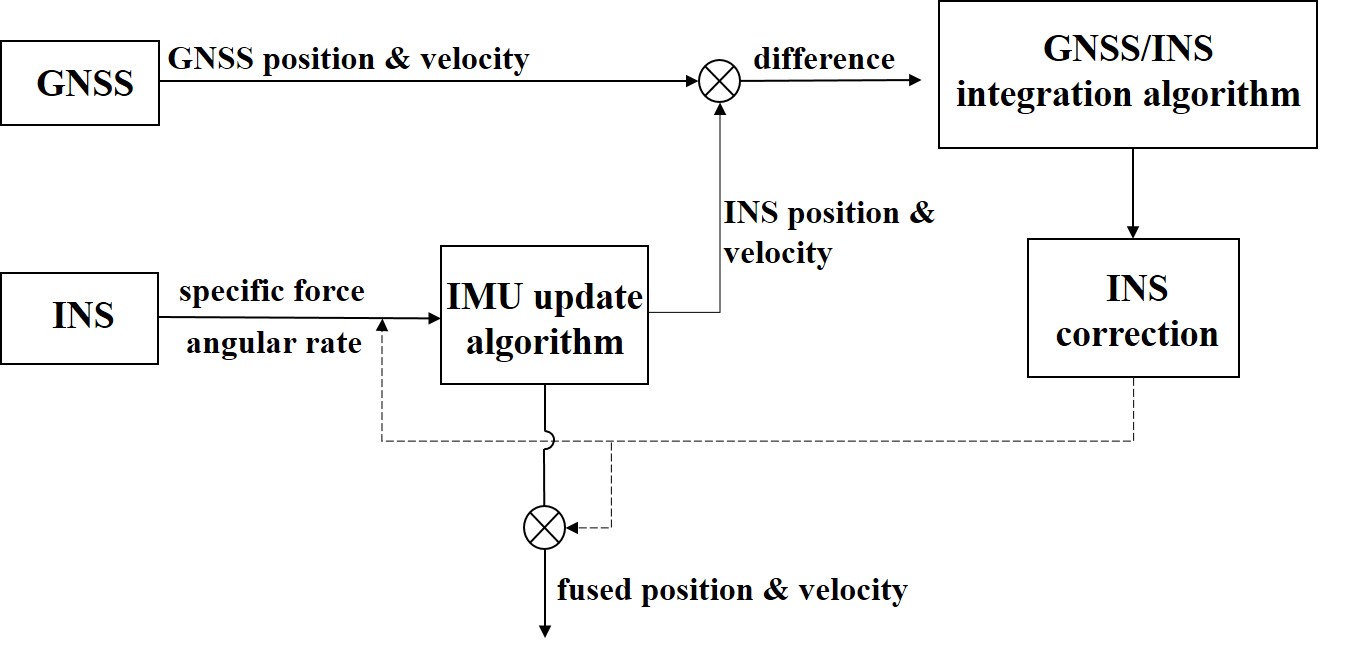}
% where an .eps filename suffix will be assumed under latex, 
% and a .pdf suffix will be assumed for pdflatex; or what has been declared
% via \DeclareGraphicsExtensions.
\caption{The architecture of the GNSS/INS integration for vehicle self-positioning.}
\label{fig2}
\end{figure}

For simplicity, we apply the GNSS/INS loosely coupled fusion system for vehicle self-positioning. The differences between GNSS-measured position and velocity and those derived from INS are used to estimate INS errors. The estimated navigation errors, including the position error, the velocity error, and the angle error, are introduced to refine the INS-derived solutions. Furthermore, the estimated biases of the gyroscope and accelerometer are feedback to correct raw IMU measurements. The system architecture is shown in Fig.\ref{fig2}.

%Fig.\ref{fig2} illustrates the system architecture of the GNSS/INS loosely coulped (LC) fusion system. 

The state prediction equation is as follows,
\begin{equation}
{\bf{x}}(k) = {\bf{F}}(k - 1){\bf{x}}(k - 1) + {\bf{w}}(k),
\label{eq1}
\end{equation}
with
\begin{equation}
\begin{split}
{\bf{x}}(k) &= [\delta {\phi _{\rm{R}}},\delta {\phi _{\rm{P}}},\delta {\phi _Y},\delta {\rm{v_N}},\delta {\rm{v_E}},\delta {\rm{v_D}},\delta L,\delta \lambda ,\delta h,\\
&{\varepsilon _x},{\varepsilon _y},{\varepsilon _z},{\Delta _x},{\Delta _y},{\Delta _z}]^{\rm{{\rm T}}}
\label{eq2}
\end{split}
\end{equation}
where ${\bf{x}}(k)$ is the state vector with fifteen state variables, $\delta {\phi _{\rm{R}}}$, $\delta {\phi _{\rm{P}}}$, and $\delta {\phi _{\rm{Y}}}$ are roll, pitch, and yaw angle errors, respectively;$\delta {v_{\rm{N}}}$, $\delta {v_{\rm{E}}}$, and $\delta {v_{\rm{D}}}$ are velocity errors from the direction of north, east, and down, respectively; $\delta L$, $\delta \lambda$, and $\delta h$ are position errors of latitude, longitude, and height in the geodetic coordinate system, respectively. ${\varepsilon _x}$, ${\varepsilon _y}$, and ${\varepsilon _z}$ denote constant gyro drifts, and ${\Delta _x}$ , ${\Delta _y}$ and ${\Delta _z}$ denote constant accelerometer drifts. Besides, the process noise vector ${\bf{w}}(k)$ is a Gaussian random vector having covariance matrix ${\bf{Q}}(k)$. ${\bf{F}}(k - 1)$ is the state transition matrix, which connects the INS solution and IMU error sources. More details about the state transition matrix can be found in ref \cite{29}\cite{30}, which is well-known in inertial theory.

The velocity and position obtained from the INS solution and GNSS receiver are defined as ${{\bf{v}}_{{\rm{INS}}}}$, ${{\bf{r}}_{{\rm{INS}}}}$, ${{\bf{v}}_{{\rm{GNSS}}}}$ and ${{\bf{r}}_{{\rm{GNSS}}}}$ respectively. 
\begin{equation}
{{\bf{v}}_{{\rm{INS}}}} = {\bf{v}} + \delta {{\bf{v}}_{{\rm{INS}}}},
\label{eq3}
\end{equation}
\begin{equation}
{{\bf{r}}_{{\rm{INS}}}} = {\bf{r}} + \delta {{\bf{r}}_{{\rm{INS}}}},
\end{equation}
\begin{equation}
{{\bf{v}}_{{\rm{GNSS}}}} = {\bf{v}}{\rm{ - }}\delta {{\bf{v}}_{{\rm{GNSS}}}},
\end{equation}
\begin{equation}
{{\bf{r}}_{{\rm{GNSS}}}} = {\bf{r}} - \delta {{\bf{r}}_{{\rm{GNSS}}}},
\label{eq-rgnss}
\end{equation}
with
\begin{equation*}
\delta {{\bf{v}}_{{\rm{INS}}}}{\rm{ = [}}\begin{array}{*{20}{c}}
{\delta {v_{\rm{N}}}}&{\begin{array}{*{20}{c}}
{\delta {v_{\rm{E}}}}&{\delta {v_{\rm{D}}}}
\end{array}}
\end{array}{{\rm{]}}^{\rm{T}}},
\end{equation*}
\begin{equation*}
\delta {{\bf{r}}_{{\rm{INS}}}}{\rm{ = [}}\begin{array}{*{20}{c}}
{\delta L}&{\delta \lambda }&{\delta h}
\end{array}{{\rm{]}}^{\rm{T}}},
\label{eq8}
\end{equation*}
where ${\bf{v}}$ and ${\bf{r}}$ represent the true velocity and the position in the geodetic coordinate system, respectively; $\delta {{\bf{v}}_{{\rm{GNSS}}}}$ and $\delta {{\bf{r}}_{{\rm{GNSS}}}}$ represent the velocity and positions of the GNSS receiver, respectively.  
Consequently, the observation equation can be formulated by 
\begin{equation}
\begin{aligned}
{\bf{z}}(k) &= \left[ {\begin{array}{*{20}{c}}
{{{\bf{v}}_{{\rm{INS}}}}{\rm{ - }}{{\bf{v}}_{{\rm{GNSS}}}}}\\
{{{\bf{r}}_{{\rm{INS}}}}{\rm{ - }}{{\bf{r}}_{{\rm{GNSS}}}}}
\end{array}} \right]{\rm{ = }}\left[ {\begin{array}{*{20}{c}}
{\delta {{\bf{v}}_{{\rm{INS}}}}{\rm{ + }}\delta {{\bf{v}}_{{\rm{GNSS}}}}}\\
{\delta {{\bf{r}}_{{\rm{INS}}}}{\rm{ + }}\delta {{\bf{r}}_{{\rm{GNSS}}}}}
\end{array}} \right]\\
& ={{\bf{H}}_{{\rm{sp}}}}(k){\bf{x}}(k){\rm{ + }}{{\bf{v}}_{{\rm{sp}}}}\left( k \right),
\label{eq9}
\end{aligned}
\end{equation}
with
\begin{equation*}
{{\bf{H}}_{{\rm{sp}}}}(k){\rm{ = }}\left[ {\begin{array}{*{20}{c}}
{{{\bf{0}}_{3 \times 3}},{\mathop{\rm diag}\nolimits} [\begin{array}{*{20}{c}}
1&1&1
\end{array}],{{\bf{0}}_{3 \times 3}},{{\bf{0}}_{3 \times 6}}}\\
{{{\bf{0}}_{3 \times 3}},{{\bf{0}}_{3 \times 3}},{\mathop{\rm diag}\nolimits} [\begin{array}{*{20}{c}}
1&1&{ - 1}
\end{array}],{{\bf{0}}_{3 \times 6}}}
\end{array}} \right],
\end{equation*}
$$ {{\bf{v}}_{{\rm{sp}}}}\left( k \right) = \left[ {\begin{array}{*{20}{c}}
{\delta {{\bf{v}}_{{\rm{GNSS}}}}}\\
{\delta {{\bf{r}}_{{\rm{GNSS}}}}}
\end{array}} \right]$$
where the subscript 'sp' implies the 'self-positioning', the measurement noise ${{\bf{v}}_{{\rm{sp}}}}\left( k \right)$ is assumed following zero-mean Gaussian distribution with the covariance matrix ${\bf{R}}(k)$.
Eq.\eqref{eq1} and Eq. \eqref{eq9} constitute a typical linear system. With a Kalman filter, we can estimate the real-time vehicular position.

\subsection{Stop line aided GNSS/INS integration method}

The first stopped vehicle, i.e., the first vehicle waiting for the green light in an intersection scenario, is usually next to the stop line. It is not too close or too far from the stop line and maintains a safe distance to pedestrians. Generally, vehicles usually travel along the lane central area. Therefore, we assume that the distance from the head of the vehicle to the stop line, which is denoted by ${d_{\rm{s}}}$, and the distance from longitudinal median plane of the vehicle to the left lane line, which is denoted by ${d_{\rm{l}}}$, follow Gaussian distribution, respectively. We can collect data of first stopped vehicles from which we can estimate the statistical feature of the two distances, i.e., ${d_{\rm{s}}} \sim  \mathcal{N}{\rm{(}}{m_{{\rm{xb}}}}{\rm{,}}{\sigma^2_{{\rm{xb}}}}{\rm{)}}$ and ${d_{\rm{l}}} \sim \mathcal{N} {\rm{(}}{m_{{\rm{yb}}}}{\rm{,}}{\sigma^2_{{\rm{yb}}}}{\rm{)}}$. 

\begin{figure}[!t]
\centering
\includegraphics[width=6.5cm]{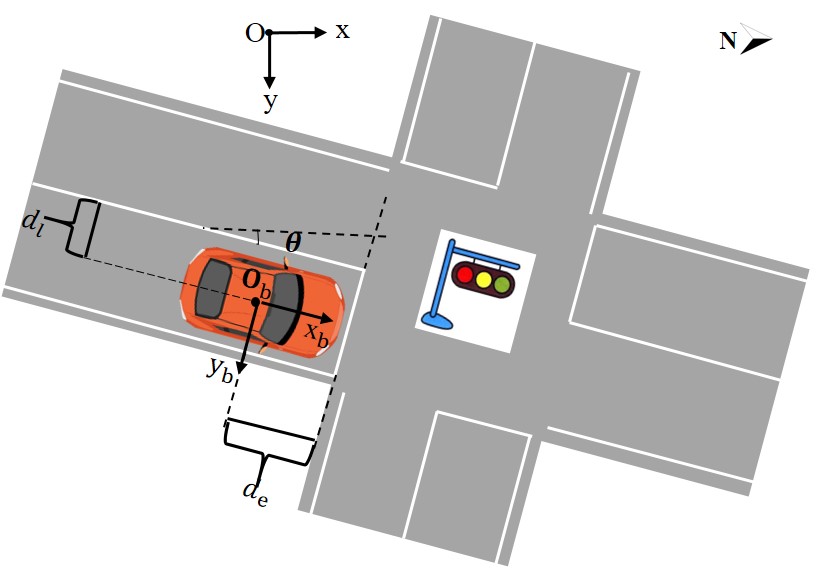}
\caption{Illustration of the stop line aided positioning for the first stopped vehicle. }
\label{fig4}
\end{figure}

No additional sensor is required to estimate the two distances, which is a great merit of the proposed method. The distance ${d_{\rm{s}}}$ and ${d_{\rm{l}}}$ can also be obtained from visual sensors, which may provide better estimation but is out of scope of this paper.

In the local north-east-down (NED) coordinate system as shown in Fig. \ref{fig4}, the stop line and the left lane line can be formulated as following two equations, respectively. 
\begin{equation*}
{y} = {k_{\rm{s}}}{x} + {b_{\rm{s}}},
\label{eq11}
\end{equation*}
\begin{equation*}
{y} = {k_{\rm{l}}}{x} + {b_{\rm{l}}},
\label{eq12}
\end{equation*}
where the slopes and the intercepts of both lines are \textit{a prior} known.

As shown in Fig.\ref{fig4}, ${d_{\rm{e}}}$, i.e., the distance from the origin $O_{\rm{b}}$ of the vehicle body coordinate system to the stop line, is given by 
$${d_{\rm{e}}}{\rm{ = }}{d_{\rm{s}}}{\rm{ + }}{l_0}$$
where ${l_0}$ represents the distance between the origin $O_{\rm b}$ and the vehicle head. 
In this paper, stop line based positioning is to estimate the position of the origin $O_{\rm b}$ in the local NED coordinate system. We denote the results of the stop line based positioning by 
$({p_{{\rm{N\_sl}}}}, {p_{{\rm{E\_sl}}}})$. Considering the equation for the distance from point to line, we can obtain
\begin{equation}
{d_e}=\frac{{\left| {{k_s}{p_{{\rm{N\_sl}}}}{\rm{ - }}{p_{{\rm{E\_sl}}}} + {b_s}} \right|}}{{\sqrt {{k_s}^2 + 1} }},
\label{eq13}
\end{equation}
\begin{equation}
{d_l}=\frac{{\left| {{k_l}{p_{{\rm{N\_sl}}}}{\rm{ - }}{p_{{\rm{E\_sl}}}} + {b_l}} \right|}}{{\sqrt {{k_l}^2 + 1} }}.
\label{eq14}
\end{equation}
From the above two equations, we can directly solve the vehicle position  $({p_{{\rm{N\_sl}}}}, {p_{{\rm{E\_sl}}}})$.

Fusing with the observations of INS, we can achieve the following observation equation in the local NED coordinate system: 
\begin{equation}
\begin{aligned}
{{\bf{z}}_{{\rm{sl}}}}(k) &= \left[ {\begin{array}{*{20}{c}}
{{p_{{\rm{N\_INS}}}} - {p_{{\rm{N\_sl}}}}}\\
{{p_{{\rm{E\_INS}}}} - {p_{{\rm{E\_sl}}}}}
\end{array}} \right] = \left[ {\begin{array}{*{20}{c}}
{\delta {p_{{\rm{N\_INS}}}}{\rm{ - }}\delta {p_{{\rm{N\_sl}}}}}\\
{\delta {p_{{\rm{E\_INS}}}}{\rm{ - }}\delta {p_{{\rm{E\_sl}}}}}
\end{array}} \right]\\
& ={{\bf{H}}_{{\rm{sl}}}}(k){\bf{x}}(k){\rm{ + }}{{\bf{v}}_{{\rm{sl}}}}\left( k \right)
\end{aligned}
\end{equation}
with $${{\bf{H}}_{{\rm{sl}}}}(k){\rm{ = }}\left[ {\begin{array}{*{20}{c}}
{{{\bf{0}}_{1 \times 6}},{R_{\rm{M}}}{\rm{ + }}h,{{\bf{0}}_{1 \times 8}}}\\
{{{\bf{0}}_{1 \times 7}},{R_{\rm{N}}},{{\bf{0}}_{1 \times 7}}}
\end{array}} \right],$$

$${{\bf{v}}_{{\rm{sl}}}}\left( k \right)= \left[ {\begin{array}{*{20}{c}}
-\delta {p_{{\rm{N\_sl}}}}\\
-\delta {p_{{\rm{E\_sl}}}}
\end{array}} \right]$$
where ${p_{{\rm{N\_INS}}}}$ and ${p_{{\rm{E\_INS}}}}$ represent the INS-derived positioning results; $\delta {p_{{\rm{N\_INS}}}}$ and $\delta {p_{{\rm{E\_INS}}}}$ denote the INS-derived positioning errors; $\delta {p_{{\rm{N\_sl}}}}$ and $\delta {p_{{\rm{E\_sl}}}}$ are the errors of north and east position obtained from Eq. \eqref{eq13} and Eq. \eqref{eq14}. They are all formulated in the NED coordinate frame. The matrix ${{\bf{H}}_{{\rm{sl}}}}$ is used to transform the errors of the NED frame to those of the geodetic coordinate system. ${{R_{\rm{M}}}}$ and ${{R_{\rm{N}}}}$ are the long and short semi-axis radius of the Earth individually, and $h$ is the altitude of the vehicle in the geodetic coordinate system. The noise vector ${{\bf{v}}_{{\rm{sl}}}}$ is assumed following the zero-mean Gaussian distribution with the covariance matrix
\begin{equation*}
 {{\bf{\Sigma }}_{{\rm{sl}}}} = \left[ {\begin{array}{*{20}{c}}
{\sigma ({x_{\rm{N}}},{x_{\rm{N}}})}&{\sigma ({x_{\rm{N}}},{x_{\rm{E}}})}\\
{\sigma ({x_{\rm{E}}},{x_{\rm{N}}})}&{\sigma ({x_{\rm{E}}},{x_{\rm{E}}})},
\end{array}} \right]  
\end{equation*}
where $\sigma ({x_{\rm{N}}},{x_{\rm{N}}})$ and $\sigma ({x_{\rm{E}}},{x_{\rm{E}}})$ denote the error variance in the north and east direction, respectively; $\sigma ({x_{\rm{N}}},{x_{\rm{E}}}){\rm{ = }}\sigma ({x_{\rm{E}}},{x_{\rm{N}}})$ represents the error covariance in the north and east direction. ${{\bf{\Sigma }}_{{\rm{sl}}}}$ can be obtained by coordinate conversion from ${\bf{\Sigma }}$ representing the covariance matrix of the observed longitude distance ${d_{\rm{s}}}$ and lateral distance ${d_{\rm{l}}}$. Apparently, the errors in the longitudinal and lateral direction of the road are uncorrelated, but the errors are correlated in the north and east directions unless the road direction is due north or due east. Denoting the coordinate transformation matrix by ${\bf{T}}$ and the angle between the road and due north by $\theta$ as shown in Fig. \ref{fig4}, we have
\begin{equation}
{{\bf{\Sigma }}_{{\rm{sl}}}} = {\bf{T\Sigma }}{{\bf{T}}^{\rm{T}}},
\end{equation}
where 
\begin{equation*}
{\bf{T}} = \left[ {\begin{array}{*{20}{c}}
{\cos \theta }&{ - \sin \theta }\\
{\sin \theta }&{\cos \theta }
\end{array}} \right],\quad\quad
{\bf{\Sigma }}{\rm{ = }}\left[ {\begin{array}{*{20}{c}}
{{\sigma _{{\rm{xb}}}}^2}&0\\
0&{{\sigma _{{\rm{yb}}}}^2}
\end{array}} \right].
\end{equation*}

\begin{figure}[!t]
\centering
\includegraphics[scale=0.3]{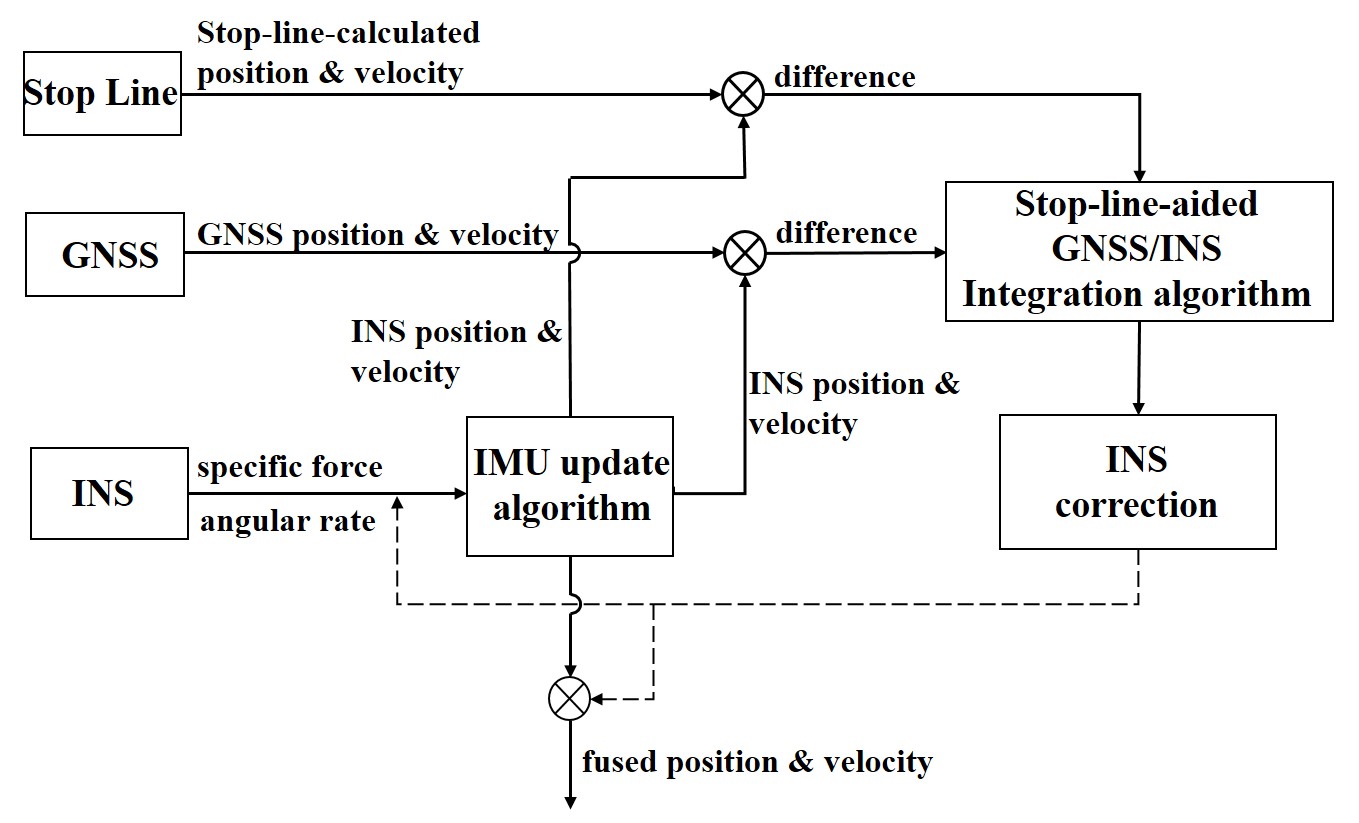}
\caption{The architecture of the stop line aided GNSS/INS integration for vehicle self-positioning.}
\label{fig3}
\end{figure}

Combining with observations from the GNSS receivers, the observation equation can be finally formulated to be 
\begin{equation}
{\bf{Z}}(k) = \left[ {\begin{array}{*{20}{c}}
{{p_{{\rm{N\_INS}}}} - {p_{{\rm{N\_sl}}}}}\\
{{p_{{\rm{E\_INS}}}} - {p_{{\rm{E\_sl}}}}}\\
{{{\bf{v}}_{{\rm{INS}}}}{\rm{ - }}{{\bf{v}}_{{\rm{GNSS}}}}}\\
{{{\bf{r}}_{{\rm{INS}}}}{\rm{ - }}{{\bf{r}}_{{\rm{GNSS}}}}}
\end{array}} \right] = {{\bf{H}}_{{\rm{sp\_sl}}}}(k){\bf{x}}(k){\rm{ + }}{{\bf{v}}_{{\rm{sp\_sl}}}}\left( k \right).
\label{eq18}
\end{equation}
Applying the Kalman filter for the linear system consisting of Eq.\eqref{eq1} and Eq.\eqref{eq18}, we can obtain the stop-line corrected vehicular position. The architecture of the stop line aided positioning enhancement can be found from Fig. \ref{fig3}.

\section{Cooperative Inertial Navigation Approach}

For connected vehicles in intersection scenarios, with the positioning improvement of the first stopped vehicle, we aim to further extend the enhancement to all vehicles in the VANET via a cooperative inertial navigation framework. 

\begin{figure}[!ht]
\centering
\includegraphics[scale=0.4]{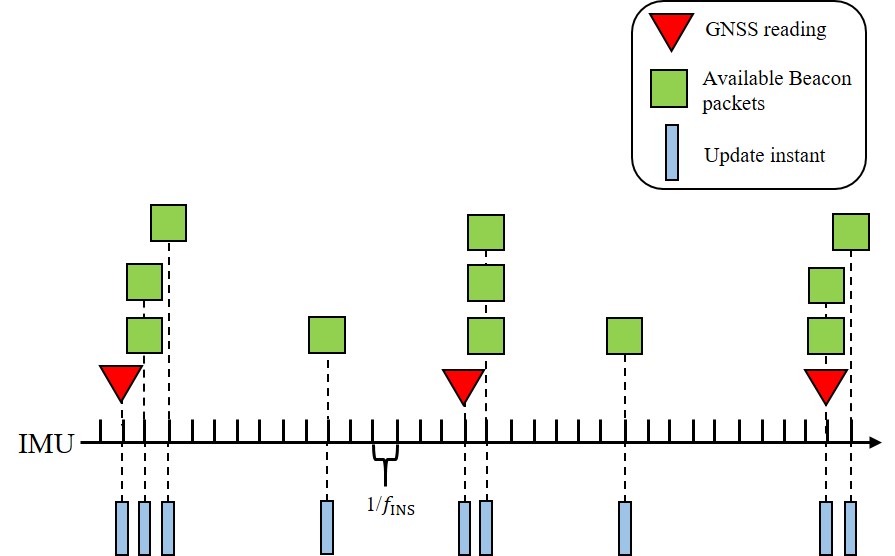}
\caption{Illustration of the observation sequences.}
\label{figRate}
\end{figure}

In such a framework, if vehicle $i$ interacts with other vehicles, we run a cooperative positioning algorithm for vehicle $i$. The elements of the state vector ${\bf x}^{(i)}(k)$ are the same as those of the state vector ${\bf x}(k)$ in Eq. \eqref{eq2}. The cooperative positioning algorithm has two stages, and the first stage is the following one-step prediction:
\begin{equation}
{\bf{x}}^{(i)}(k|k - 1) = {\bf{F}}(k\left| {k - 1} \right.){\bf{x}}^{(i)}(k{\rm{ - }}1),
\label{eqPre}
\end{equation}
\begin{equation}
{\bf{P}}^{(i)}(k|k - 1) = {\bf{F}}(k - 1){\bf{P}}^{(i)}(k - 1){\bf{F}}^{\rm T}{( k - 1 )}{\rm{ + }}{\bf{Q}}{\rm{(}}k{\rm{ - 1)}}.
\end{equation}
The arrival of new information triggers the second stage, i.e., the update stage. Except for the periodic GNSS observation, vehicle $i$ also measures the inter-node distances and receives beacon packets from neighbors both via V2V communication. Similar to the common GNSS/INS integration method, we adopt newly arrived information to correct INS errors.
However, it unnecessarily receives all above-mentioned information during a time horizon due to the working frequencies difference between the GNSS and the V2V communications. As shown in Fig.\ref{figRate}, the filter can be triggered by the reception of local GNSS readings or the received beacon packets from neighbors. Consequently, the update stage at the $k$ instant has the following three cases:
 
\begin{itemize}
\item Case1:only the local GNSS readings are available;
\item Case2:the beacon packets and the inter-node distances are available; 
\item Case3:the local GNSS readings, the beacon packets, and the inter-node distances are all available.
\end{itemize}

\subsection{Case1:Only the local GNSS readings are available}

Except for the INS observations, if only the local GNSS observations are available, it is a typical self-positioning problem. 
\subsubsection{Vehicle $i$ is the first stopped vehicle} As discussed in the last section, the INS-derived solutions can be corrected by the stop line, we use the Eq. \eqref{eq18} as the observation equation, i.e.,
\begin{equation}
{\bf{z}}(k){\rm{ = }}{{\bf{z}}^{(i)}_{{\rm{sp\_sl}}}}(k){\rm{ = }}{{\bf{H}}^{(i)}_{{\rm{sp\_sl}}}}(k){\bf{x}}(k){\rm{ + }}{{\bf{v}}^{(i)}_{{\rm{sp\_sl}}}}\left( k \right).
\label{eq23}
\end{equation}
\subsubsection{Vehicle $i$ is not the first stopped vehicle} this case is a typical GNSS/INS integrated navigation system, and the observation equation is the Eq.\eqref{eq9}, i.e.,
\begin{equation}
{\bf{z}}(k){\rm{ = }}{{\bf{z}}^{(i)}_{{\rm{sp}}}}(k){\rm{ = }}{{\bf{H}}^{(i)}_{{\rm{sp}}}}(k){\bf{x}}(k){\rm{ + }}{{\bf{v}}^{(i)}_{{\rm{sp}}}}\left( k \right).
\label{eq22}
\end{equation}

\subsection{Case2:Beacon packets and inter-node distances are available}

We denote the INS calculation and true location of vehicle $i$ by  ${\bf{p}}_{_{{\rm{INS}}}}^{(i)}(k){\rm{ = }}{\left[ {{p_{{\rm{N\_INS}}}},{p_{{\rm{E\_INS}}}}} \right]^{\rm{T}}}$ and ${{\bf{p}}^{(i)}}(k)$, respectively. Like Eq. $(\ref{eq3}) \sim \eqref{eq-rgnss}$, we have
\begin{equation}
{\bf{p}}_{_{{\rm{INS}}}}^{(i)}(k){\rm{ = }}{{\bf{p}}^{(i)}}(k){\rm{ + }}\delta {{\bf{p}}^{(i)}}(k)
\end{equation}
where $\delta {{\bf{p}}^{(i)}} = {\left[ {\delta {p_{{\rm{N\_INS}}}},\delta {p_{{\rm{E\_INS}}}}} \right]^{\rm{T}}}$ is the INS positioning error in the NED coordinate system. $\delta {{\bf{p}}^{(i)}}$ can be transformed into the geodetic coordinate system, i.e.,
\begin{equation}
\delta {{\bf{p}}^{(i)}}{\rm{ = [}}({R_{\rm{M}}}{\rm{ + }}h){\bf{x}}(7),{R_{\rm{N}}}{\bf{x}}(8)].
\label{eq25}
\end{equation}

 We assume vehicle $i$ receives beacon packets and relative distances from its $n$ neighbor vehicles, and denote the  neighborhood set by ${S_{i}}$.  In the NED coordinate frame, $\forall j \in {S_{i}},j = 1,2,...n$, we denote the self-positioning estimation and true location of vehicle $j$ by ${{\bf{\hat p}}^{(j)}}(k)$ and ${{\bf{p}}^{(j)}}(k)$, respectively. We introduce ${\tilde z_{j \to i}}$, the difference between the estimated position of vehicle $i$ and that of vehicle $j$, and the relative distance ${\hat z_{j \to i}}$ measured from the V2V communication. Then we arrive at the following V2V-based observation equation: 
\begin{equation}
\begin{split}
{\rm{z}}(k)&={\rm{z}}_{_{{\rm{v2v}}}}^{(j)}{\rm{(}}k{\rm{)}}\\
&={\tilde z_{j \to i}} - {\hat z_{j \to i}}\\
&= \left\| {{\bf{p}}_{_{{\rm{INS}}}}^{(i)}{\rm{(}}k{\rm{) - }}{{{\bf{\hat p}}}^{(j)}}(k)} \right\| - {\hat z_{j \to i}}\\
&=\left\| {[{{\bf{p}}^{{\rm{(}}i{\rm{)}}}}{\rm{(}}k{\rm{)}} + \delta {{\bf{p}}^{{\rm{(}}i{\rm{)}}}}{\rm{(}}k{\rm{)}}]{\rm{ - [}}{{\bf{p}}^{{\rm{(}}j{\rm{)}}}}{\rm{(}}k{\rm{) + }}\delta {{\bf{p}}^{{\rm{(}}j{\rm{)}}}}{\rm{(}}k{\rm{)}}]} \right\| \\
&- (\left\| {{{\bf{p}}^{{\rm{(}}i{\rm{)}}}}{\rm{(}}k{\rm{) - }}{{\bf{p}}^{{\rm{(}}j{\rm{)}}}}{\rm{(}}k{\rm{)}}} \right\| - {n_{j \to i}})\\
&={\mathop{\rm h}\nolimits} _{j \to i}^{\rm{p}}(\delta {{\bf{p}}}) + {n_{j \to i}}, 
\label{eq26}
\end{split}
\end{equation}
with  
$$\delta {\bf{p}} = [\begin{array}{*{20}{c}}
{\delta {{\bf{p}}^{{\rm{(}}i{\rm{)}}}}}&{\delta {{\bf{p}}^{{\rm{(}}j{\rm{)}}}}}
\end{array}]$$
where $\delta {{\bf{p}}^{{\rm{(}}j{\rm{)}}}}{\rm{(}}k{\rm{) = }}{{\bf{\hat p}}^{(j)}}(k){\rm{ - }}{{\bf{p}}^{(j)}}(k)$ is the self-positioning error of the vehicle $j$, and ${n_{j \to i}}$ is the V2V ranging error. Since ${\mathop{\rm h}\nolimits} _{j \to i}^{\rm{p}}(\delta {\bf{p}})$ is a nonlinear function of $\delta {{\bf{p}}}$, linearizing ${\mathop{\rm h}\nolimits}_{j \to i}^{\rm{p}}(\delta {\bf{p}})$ yields

\begin{equation}
\begin{split}
{{\bf{z}}^{(j)}_{{\rm{v2v}}}}\left( k \right)&={\mathop{\rm h}\nolimits} _{j \to i}^{\rm{p}}(\delta {\bf{p}})  + {n_{j \to i}} \\
&\approx {\bf{H}}_{j \to i}^p\delta {{\bf{p}}^{(i)}}{\rm{(}}k{\rm{) + }}{\bf{H}}_{i \to j}^{\rm{p}}\delta {{\bf{p}}^{{\rm{(}}j{\rm{)}}}}{\rm{(}}k{\rm{) + }}{n_{j \to i}} \\
&= {{\bf{H}}^{(j)}_{{\rm{v2v}}}}{\bf{x}}{\rm{(}}k{\rm{) + }}{\bf{H}}_{i \to j}^p\delta {{\bf{p}}^{{\rm{(}}j{\rm{)}}}}{\rm{(}}k{\rm{) + }}{n_{j \to i}}\\
&={{\bf{H}}^{(j)}_{{\rm{v2v}}}}{\bf{x}}{\rm{(}}k{\rm{)}} + {u_{j}}{\rm{ + }}{n_{j \to i}}\\
&={{\bf{H}}^{(j)}_{{\rm{v2v}}}}{\bf{x}}{\rm{(}}k{\rm{) + }}{{\bf{v}}^{(j)}_{{\rm{v2v}}}}{\rm{(}}k{\rm{)}}
\label{eq27}
\end{split}
\end{equation}
where 
\begin{equation*}
{\bf{H}}_{j \to i}^{\rm{p}}{\rm{ =  - }}{\bf{H}}_{i \to j}^{\rm{p}}{\rm{ = }}{\left. {\frac{{\partial {\mathop{\rm h}\nolimits} _{j \to i}^{\rm{p}}(\delta {\bf{p}})}}{{\partial \delta {{\bf{p}}^{(i)}}}}} \right|_{\delta {\bf{p}} = [\delta {{\bf{p}}^{(i)}}(k\left| {k - 1} \right.),\delta {{\bf{p}}^{(j)}}(k\left| {k - 1} \right.)]}}
\end{equation*}
is the Jacobian matrix of ${\mathop{\rm h}\nolimits} _{j \to i}^{\rm{p}}(\delta {{\bf{p}}})$ at the current state prediction which can be obtained from Eq.\eqref{eqPre}, and the measurement matrix 
${{\bf{H}}^{(j)}_{{\rm{v2v}}}}{\rm{ = }}\left[ {{{\bf{0}}_{1 \times 6}},({R_{\rm{M}}}{\rm{ + }}h){\bf{H}}_{j \to i}^{\rm{p}}(1),{R_{\rm{N}}}{\bf{H}}_{j \to i}^{\rm{p}}(2),{{\bf{0}}_{1 \times 7}}} \right].$
In Eq. \eqref{eq27}, ${u_{j}} = {\bf{H}}_{i \to j}^{\rm{p}}\delta {{\bf{p}}^{{\rm{(}}j{\rm{)}}}}{\rm{(}}k{\rm{) }}$ follows a Gaussian distribution since $\delta{{\bf{p}}^{{\rm{(}}j{\rm{)}}}}{\rm{(}}k{\rm{) }}$ follows a Gaussian distribution. Denote ${u_{j}} \sim \mathcal{N}{\rm{(0,}}{r_{j}}{\rm{)}} $, we have 
\begin{equation}
\begin{split}
{r_{j}}&={\bf{H}}_{i \to j}^{\rm{p}}\mathds{E}[\delta {{\bf{p}}^{{\rm{(}}j{\rm{)}}}}{\rm{(}}k{\rm{)(}}\delta {{\bf{p}}^{{\rm{(}}j{\rm{)}}}}{\rm{(}}k{\rm{)}}{{\rm{)}}^{\rm{T}}}]{({\bf{H}}_{i \to j}^{\rm{p}})^{\rm{T}}}\\
&={\bf{H}}_{i \to j}^{\rm{p}}{{\bf{P}}^{(j)}}(k){({\bf{H}}_{i \to j}^{\rm{p}})^{\rm{T}}}.
\end{split}
\end{equation}
Here ${{\bf{P}}^{(j)}}(k)=\mathds{E}[\delta {{\bf{p}}^{{\rm{(}}j{\rm{)}}}}{\rm{(}}k{\rm{)(}}\delta {{\bf{p}}^{{\rm{(}}j{\rm{)}}}}{\rm{(}}k{\rm{)}}{{\rm{)}}^{\rm{T}}}]$ is the covariance matrix of the self-positioning result of vehicle $j$.  The measurement noise of the relative distance between two vehicles is also assumed as a Gaussian distribution, i.e., ${n_{j \to i}} \sim \mathcal{N} {\rm{(0,}}\sigma _{{\rm{V2V}}}^2{\rm{)}}$, and the variance $\sigma^2_{\rm{V2V}}$ can be estimated from  \textit{a prior} observed data-sets for V2V-based distance measurements.

If vehicle $j$ is the first stopped vehicle, the position estimation will be corrected with the aid of the stop line, and consequently has a smaller covariance ${{\bf{P}}^{(j)}}(k)$, which leads to a smaller $r_j$. In that case, vehicle $j$ can be viewed as a temporary anchor such that its neighbors can apply the proposed cooperative positioning framework to achieve better positioning results.

If vehicle $i$ can receive beacon packets and relative distances from all its $n$ neighbor nodes, we generalize the above mentioned vehicle $j$ to all vehicles in $S_i$, and arrive at the observation equation (\ref{eq209}).

\begin{figure*}[ht]
\begin{equation}
{\bf{z}}\left( k \right) = \left[ {\begin{array}{*{20}{c}}
{{\rm{z}}_{_{{\rm{v2v}}}}^{(1)}\left( k \right)}\\
{\begin{array}{*{20}{c}}
{\begin{array}{*{20}{c}}
{\begin{array}{*{20}{c}}
{{\rm{z}}_{_{{\rm{v2v}}}}^{(2)}\left( k \right)}\\
{...}
\end{array}}
\end{array}}\\
{{\rm{z}}_{_{{\rm{v2v}}}}^{(n)}\left( k \right)}
\end{array}}
\end{array}} \right]{\rm{ = }}\left[ {\begin{array}{*{20}{c}}
{{\bf{H}}_{_{{\rm{v2v}}}}^{(1)}\left( k \right)}\\
{\begin{array}{*{20}{c}}
{\begin{array}{*{20}{c}}
{\begin{array}{*{20}{c}}
{{\bf{H}}_{_{{\rm{v2v}}}}^{(2)}\left( k \right)}\\
{...}
\end{array}}
\end{array}}\\
{{\bf{H}}_{_{{\rm{v2v}}}}^{(n)}\left( k \right)}
\end{array}}
\end{array}} \right]{\bf{x}}(k)){\rm{ + }}\left[ {\begin{array}{*{20}{c}}
{{\bf{v}}_{_{{\rm{v2v}}}}^{(1)}\left( k \right)}\\
{\begin{array}{*{20}{c}}
{\begin{array}{*{20}{c}}
{\begin{array}{*{20}{c}}
{{\bf{v}}_{_{{\rm{v2v}}}}^{(2)}\left( k \right)}\\
{...}
\end{array}}
\end{array}}\\
{{\bf{v}}_{_{{\rm{v2v}}}}^{(n)}\left( k \right)}
\end{array}}
\end{array}} \right]
\label{eq209}
\end{equation}
\end{figure*}

\begin{figure*}[ht]
\begin{equation}
{\bf{z}}\left( k \right) = \left[ {\begin{array}{*{20}{c}}
{\begin{array}{*{20}{c}}
{{{\bf{z}}^{(i)}_{{\rm{sp\_sl}}}}\left( k \right)\;\;\mathrm{or}\;\;{{\bf{z}}^{(i)}_{{\rm{sp}}}}\left( k \right)}\\
{{{\bf{z}}^{(1)}_{{\rm{v2v}}}}\left( k \right)}
\end{array}}\\
{\begin{array}{*{20}{c}}
{\begin{array}{*{20}{c}}
{\begin{array}{*{20}{c}}
{{{\bf{z}}^{(2)}_{{\rm{v2v}}}}\left( k \right)}\\
{...}
\end{array}}
\end{array}}\\
{{{\bf{z}}^{(n)}_{{\rm{v2v}}}}\left( k \right)}
\end{array}}
\end{array}} \right]{\rm{ = }}\left[ {\begin{array}{*{20}{c}}
{\begin{array}{*{20}{c}}
{{{\bf{H}}^{(i)}_{{\rm{sp\_sl}}}}\;\;\mathrm{or}\;\;{{\bf{H}}^{(i)}_{{\rm{sp}}}}}\\
{{{\bf{H}}^{(1)}_{{\rm{v2v}}}}}
\end{array}}\\
{\begin{array}{*{20}{c}}
{\begin{array}{*{20}{c}}
{\begin{array}{*{20}{c}}
{{{\bf{H}}^{(2)}_{{\rm{v2v}}}}}\\
{...}
\end{array}}
\end{array}}\\
{{{\bf{H}}^{(n)}_{{\rm{v2v}}}}}
\end{array}}
\end{array}} \right]{\bf{x}}(k){\rm{ + }}\left[ {\begin{array}{*{20}{c}}
{\begin{array}{*{20}{c}}
{{{\bf{v}}^{(i)}_{{\rm{sp\_sl}}}}\left( k \right)\;\;\mathrm{or}\;\;{{\bf{v}}^{(i)}_{{\rm{sp}}}}\left( k \right)}\\
{{{\bf{v}}^{(1)}_{{\rm{v2v}}}}\left( k \right)}
\end{array}}\\
{\begin{array}{*{20}{c}}
{\begin{array}{*{20}{c}}
{\begin{array}{*{20}{c}}
{{{\bf{v}}^{(2)}_{{\rm{v2v}}}}\left( k \right)}\\
{...}
\end{array}}
\end{array}}\\
{{{\bf{v}}^{(n)}_{{\rm{v2v}}}}\left( k \right)}
\end{array}}
\end{array}} \right]
\label{eq29}
\end{equation}
\end{figure*}

\begin{figure}[!ht]
\centering
\includegraphics[scale=0.32]{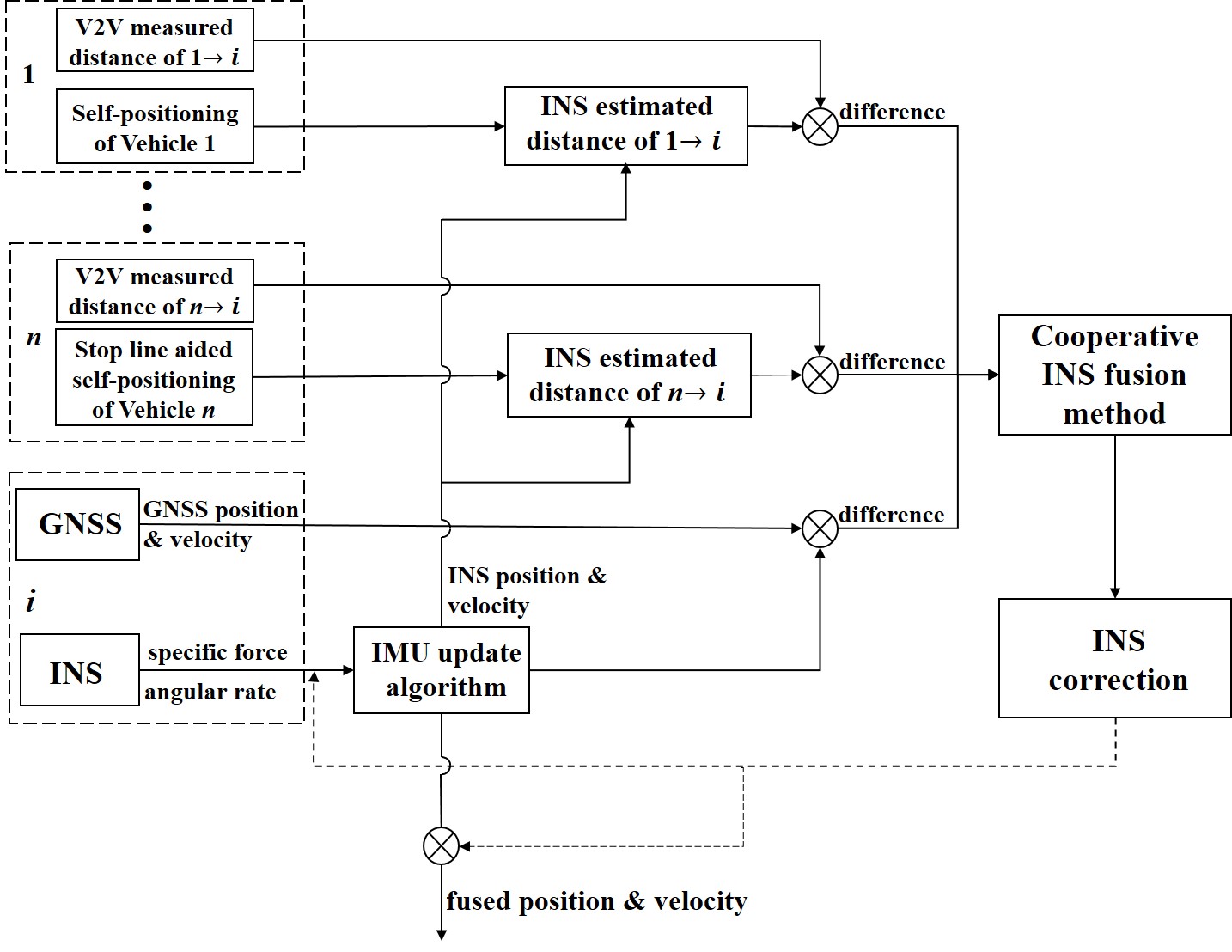}
\caption{The stop line aided cooperative positioning framework.}
\label{fig5}
\end{figure}

\subsection{Case3: local GNSS readings, beacon packets, and inter-node distances are all available}
If the local GNSS readings, beacon packets, and inter-node distances are all available,  adding the observation equation of self-positioning to Eq.\eqref{eq209} yields the observation equation \eqref{eq29}. The observation equation of self-positioning has two cases: if vehicle $i$ is the first-stopped vehicle, the observation is the ${{\bf{z}}^{(i)}_{{\rm{sp\_sl}}}}\left( k \right)$ in Eq. \eqref{eq23}; otherwise the self-positioning observation is the ${{{\bf{z}}^{(i)}_{{\rm{sp}}}}\left( k \right)}$ in Eq. \eqref{eq22}.

\subsection{Measurement update of the proposed stop line aided cooperative positioning framework}

The measurement data sequence is shown in Fig. \ref{figRate}, and the stop line aided cooperative positioning framework is shown in Fig.\ref{fig5}.  Based on the types of the received observations, we select the corresponding observation equation from Eq. \eqref{eq23}, \eqref{eq22}, \eqref{eq209}, \eqref{eq29}, respectively. For simplicity, we describe the four observation equations to be the following general form:
\begin{equation*}
    \mathbf{z}(k)=\mathbf{H}(k)\mathbf{x}(k)+\mathbf{v}(k)
\end{equation*}
which is actually one of the above-mentioned four equations.

With the observation equations, we can obtain the Kalman gain
\begin{equation}
\begin{split}
{{\bf{K}}^{(i)}}(k) &= {{\bf{P}}^{(i)}}(k|k - 1){\bf{H}}{(k)^{\rm{T}}}[{\bf{H}}(k){{\bf{P}}^{(i)}}(k|k - 1){\bf{H}}{(k)^{\rm{T}}}\\
 &+ {{\bf{R}}^{(i)}}(k){]^{ - 1}},
\end{split}
\end{equation}
the \textit{a posterior} state estimation
\begin{equation}
{\bf{\hat x}}^{(i)}(k) = {\bf{x}}^{(i)}(k|k - 1) + {\bf{K}}^{(i)}(k)\left[ {{\bf{z}}\left( k \right){\rm{ - }}{\bf{H}}(k){\bf{x}}^{(i)}(k|k - 1)} \right],
\end{equation}
and the corresponding covariance matrix 
\begin{equation}
\begin{split}
{\bf{P}}^{(i)}(k) &= ({\bf{I}} - {\bf{K}}^{(i)}(k){\bf{H}}(k)){\bf{P}}^{(i)}(k){({\bf{I}} - {\bf{K}}^{(i)}(k){\bf{H}}(k))^{\rm{T}}}\\
&+ {\bf{K}}^{(i)}(k){\bf{R}}^{(i)}(k){\bf{K}}^{(i)}{(k)^{\rm{T}}}.
\end{split}
\end{equation}

\section{Experimental Validation}
\subsection{Experimental Setup and Scenario Description}
\begin{figure}[!t]
\centering
\subfigure{               
\includegraphics[scale=0.3]{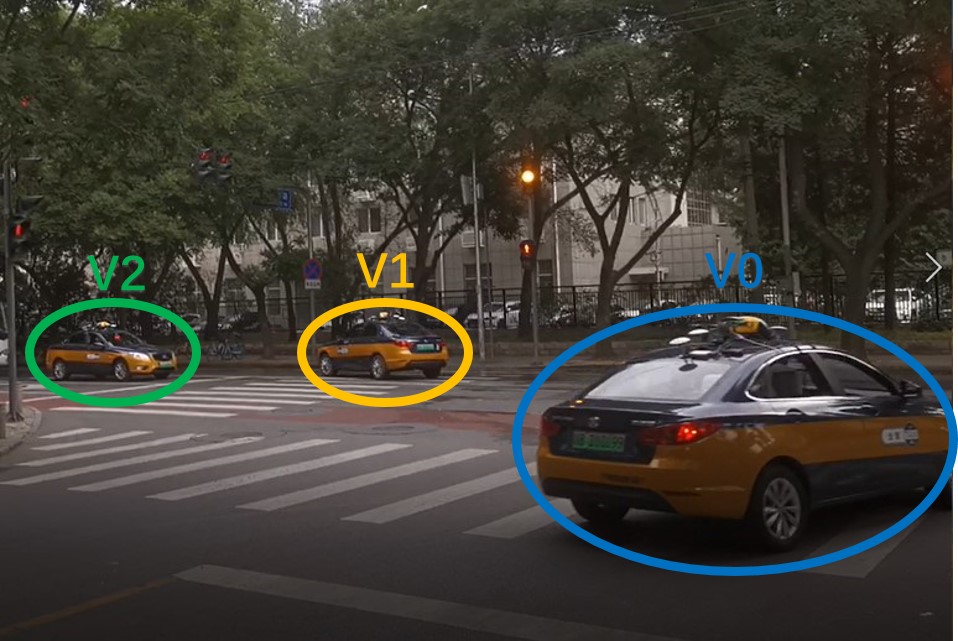}}
\subfigure{               
\includegraphics[width=3.6cm,height=3.26cm]{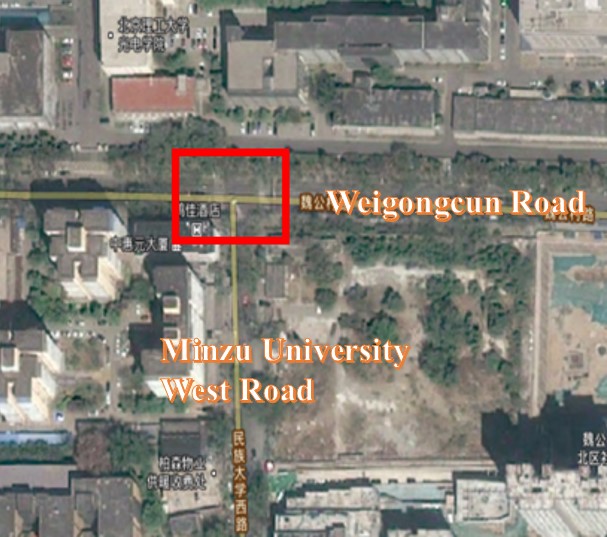}}
\caption{Experimental vehicles and the scenario setup.The red box shows the area of the intersection. Weigongcun Road features a two-way with 4-lane motor traffic and Minzu University West Road has one lane each way. }
\label{fig6}
\end{figure}

Two scenarios with three vehicles were conducted at an intersection with traffic lights in Beijing, China, as shown in Fig.\ref{fig6}. We used a high-accuracy Real-Time Kinematic GNSS/INS integrated navigation system(Novatel PwrPak7D-E1) to provide the positioning ground truth. In addition, each vehicle was equipped with a Sinan M100W GNSS navigation device, XSENS Mti630 IMU, Nooploop UWB Node, and a laptop. The M100W acted as an ordinary GNSS receiver and collected data at a frequency of 5Hz. The XSENS Mti630 IMU collected data at 100Hz. The UWB device was used for V2V communication at 5 Hz and inter-vehicle ranging at 100 Hz. 

We compare the following four positioning approaches to show the effectiveness of the proposed framework:

(1)Self-positioning (SP): GNSS/IMU loosely coupled fusion for vehicle self-positioning.

(2)Stop line aided self-positioning (SL-SP): we apply the stop line as a measurement for the first stopped vehicle and fuse the measurement with the GNSS/IMU positioning system. 

(3)Cooperative positioning (CP): the local IMU and GNSS observation, position-related information from other vehicles, and inter-vehicle ranging are applied for localization.

(4)Stop line aided cooperative positioning (SL-CP): the proposed framework.

\subsection{Experimental results for scenario 1}

This scenario shows the basic idea of the stop-line aided self-positioning enhancement and how the improvement extends to all connected vehicles. In this scenario, two vehicles travel straight from south to north on Minzu University West Road, and one vehicle turns right into Minzu University West Road from Weigongcun Road.  

\begin{figure}[]
\centering
\subfigure[the $t_0$ instant]{               
\includegraphics[width=3cm,height=3.5cm]{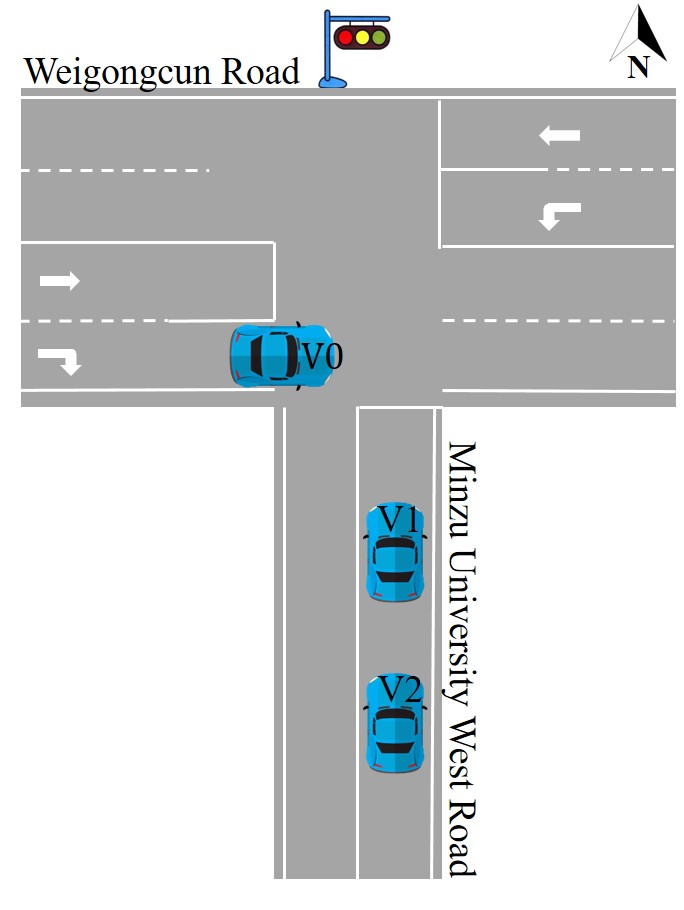}}
\hspace{0in}
\subfigure[the ${t_{s1}}$ instant]{
\includegraphics[width=3cm,height=3.5cm]{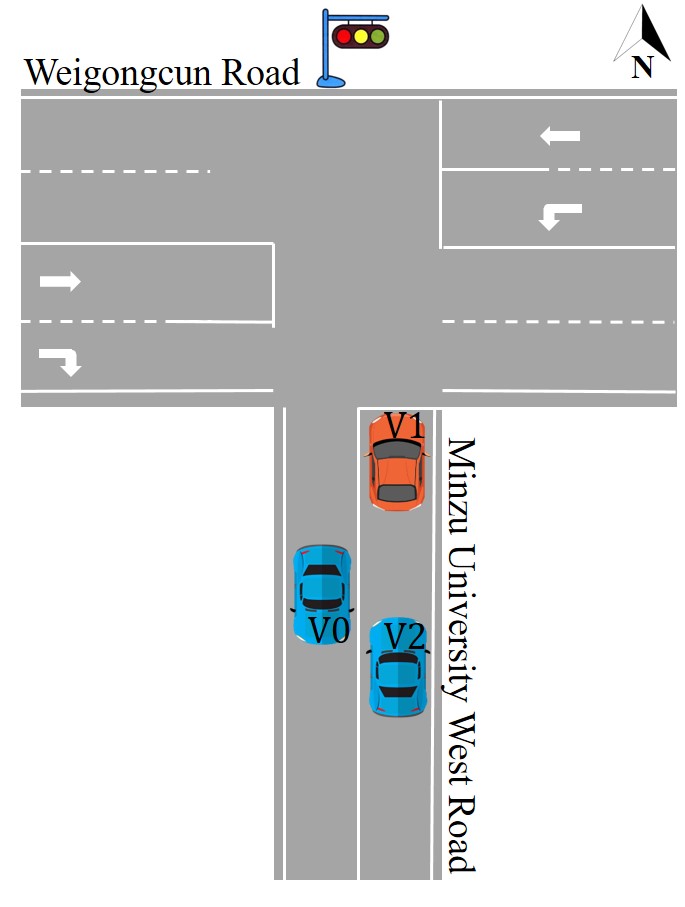}}

\subfigure[the ${t_{s2}}$ instant]{               
\includegraphics[width=3cm,height=3.5cm]{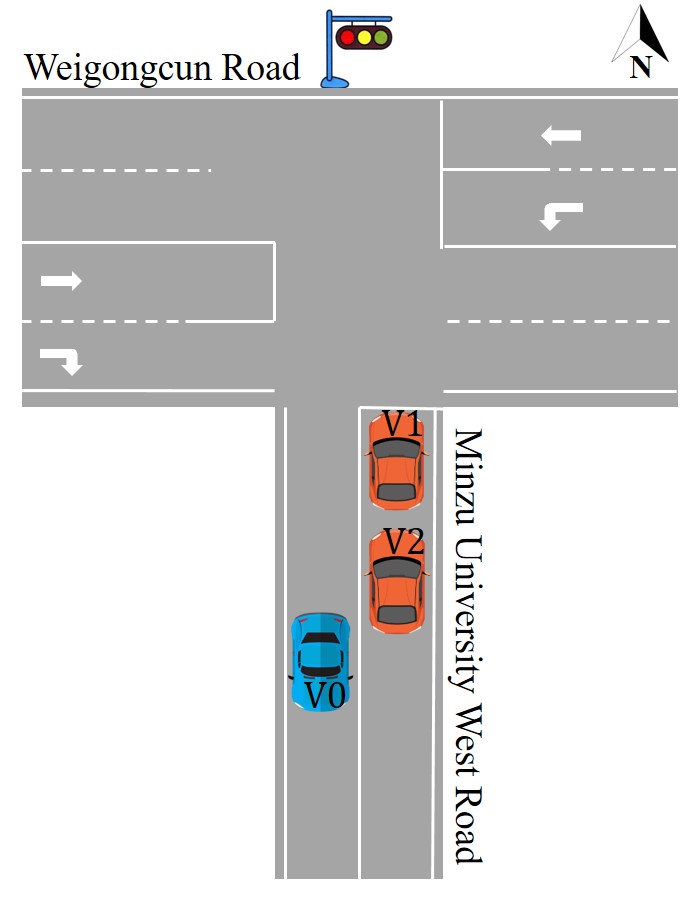}}
\hspace{0in}
\subfigure[the $t_e$ instant]{
\includegraphics[width=3cm,height=3.5cm]{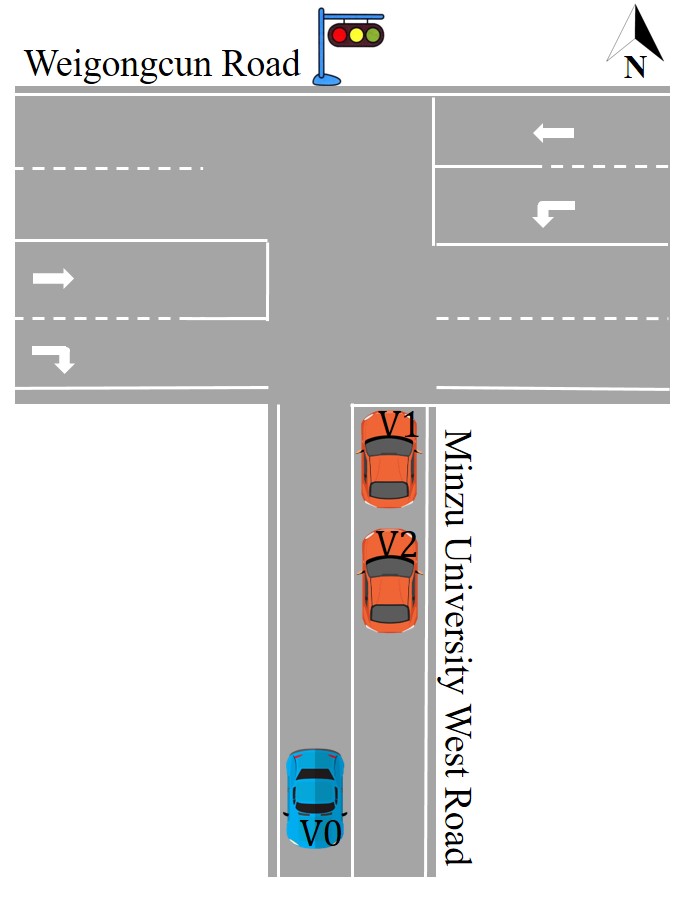}}
\caption{The key scenes of the scenario 1. Orange represents the vehicle is stopped. Blue represents the vehicle is moving.}
\label{fig8}
\end{figure}

Fig.\ref{fig8} provides the key scenes of this scenario. Fig.\ref{fig8}(a) shows that at the ${t_0}$ instant, vehicle V0 was running on the Weigongcun Road and ready to turn right into the Minzu University West Road. Vehicle V1 and V2 were both running on the Minzu University West Road.
Fig.\ref{fig8}(b) shows that at the ${t_{s1}}$ instant, vehicle V1 became the first stopped vehicle while vehicles V0 and V2 were still running. At ${t_{s2}}$, as shown in Fig.\ref{fig8}(c), both V1 and V2 were waiting for the green light while V0 was still running. At ${t_e}$, as shown in Fig.\ref{fig8}(d), both V1 and V2 were still waiting for the green light while V0 was far away from the intersection. In this scenario, the three vehicles communicated via the UWB sensors. After the ${t_e}$ instant, the communications between V0 and the other two vehicles might not be stable due to the limitation of communication distance.

\begin{figure*}[]  %全栏显示
\centering
\subfigure[V0]{               
\includegraphics[scale=0.25]{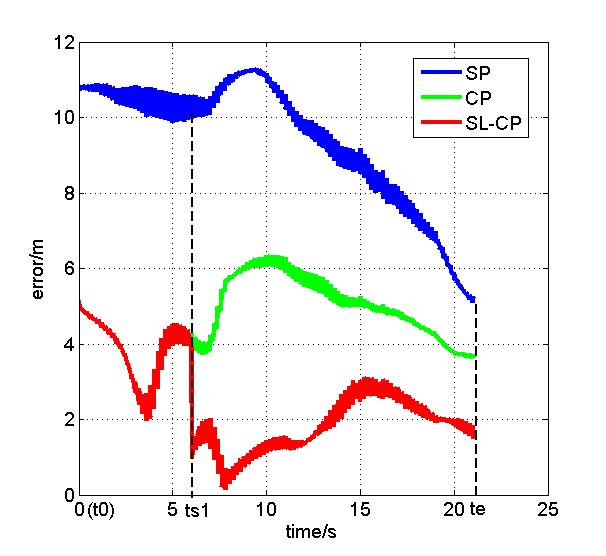}}
\hspace{0in}
\subfigure[V1]{
\includegraphics[scale=0.25]{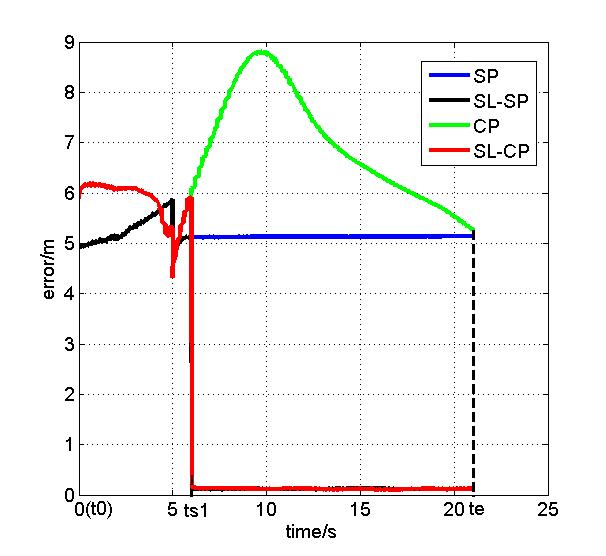}}
\hspace{0in}
\subfigure[V2]{
\includegraphics[scale=0.25]{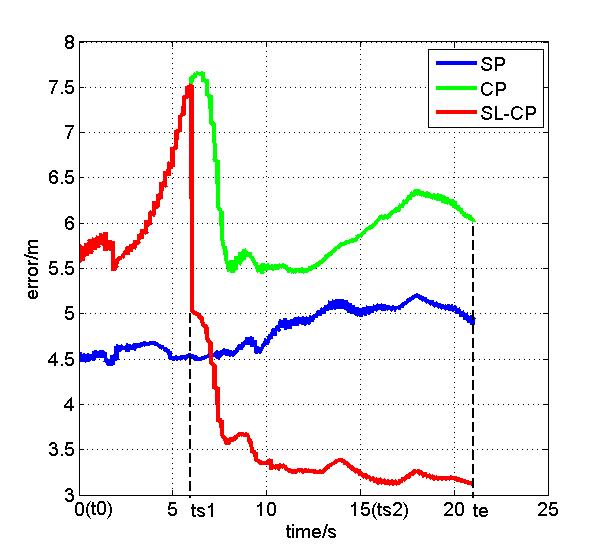}}
\caption{The positioning errors of the three vehicles in scenario 1.}
\label{fig9}
\end{figure*}

Positioning errors of the three vehicles using the four approaches as mentioned above are shown in Fig.\ref{fig9}. There was no stopped vehicle during ${t_0} \sim {t_{s1}}$. Hence, the results of SL-CP were the same as those of the cooperative positioning (CP) method during ${t_0} \sim {t_{s1}}$. Compared with self-positioning, using SL-CP or CP greatly reduced the positioning error of V0 while sightly increasing the errors of V1 and V2. The reason is that the self-positioning error of V1 and V2 were both much smaller than that of V0. A vehicle with low positioning performance can greatly benefit from the cooperative positioning strategy.  

\begin{table}[htbp]
\centering
\caption{POSITIONING ERROR (RMSE) during ${t_{s1}} \sim {t_e}$}
\begin{tabular}{p{1.5cm}p{1.5cm}p{1.5cm}p{1.5cm}}%l=left, r=right,c=center分别代表左对齐，右对齐和居中，字母的个数代表列数
\hline
 &V0 &V1 &V2\\ \hline
SP &9.15 m &5.18 m &4.91 m\\ 
SL-SP &/ m  &0.24 m &/ m\\
CP  &5.10 m &7.10 m &6.02 m \\ 
SL-CP  &1.93 m &0.25 m &3.84 m\\ \hline
\end{tabular}
\end{table}

%\begin{table}[htbp]
%\centering
%\caption{POSITIONING ERROR (RMSE) OF V0 during ${t_{s1}} \sim {t_e}$}
%\begin{tabular}{cccc}%l=left, r=right,c=center分别代表左对齐，右对齐和居中，字母的个数代表列数
%\hline
% &Error in north &Error in east &Positioning Error\\ \hline
%SP &9.14 m &0.45 m &9.15 m\\ 
%CP &4.98 m &1.12 m &5.10 m\\ 
%SL-CP &1.11 m &1.59 m &1.93 m\\ \hline
%\end{tabular}
%\end{table}

%\begin{table}[htbp]
%\centering
%\caption{POSITIONING ERROR (RMSE) OF V1 during ${t_{s1}} \sim {t_e}$}
%\begin{tabular}{cccc}%l=left, %r=right,c=center分别代表左对齐，右对齐和居中，字母的个数代表列数
%\hline
% &Error in north &Error in east &Positioning Error\\ \hline
%SP &3.50 m &4.07 m &5.37 m\\ 
%SL-SP &0.21 m &0.38 m &0.43 m\\ 
%CP &5.22 m &5.09 m &7.29 m\\ 
%SL-CP &0.33 m &0.37 m &0.50 m\\ \hline
%\end{tabular}
%\end{table}

%\begin{table}[htbp]
%\centering
%\caption{POSITIONING ERROR (RMSE) OF V2 during ${t_{s1}} \sim {t_e}$}
%\begin{tabular}{cccc}%l=left, %r=right,c=center分别代表左对齐，右对齐和居中，字母的个数代表列数
%\hline
% &Error in north &Error in east &Positioning Error\\ \hline
%SP &3.56 m &3.38 m &4.91 m\\ 
%CP &4.20 m &4.32 m &6.02 m\\ 
%SL-CP &0.94 m &3.74 m &3.84 m\\ \hline
%\end{tabular}
%\end{table}

As shown in Fig.\ref{fig8}(b), V1 was the first vehicle stopped due to the red light. It was motionless during ${t_{s1}} \sim {t_e}$. TABLE I shows the positioning errors, i.e., the root mean square error (RMSE), of the three vehicles during this period, respectively. Fig.\ref{fig9}(b) shows that at the ${t_{s1}}$ instant, the positioning error of V1 using the SL-SP method droped from about 5 m to less than 0.5 m. As can be seen from TABLE I, with the help of the stop line information, during ${t_{s1}} \sim {t_e}$, SL-SP reduced the RMSE of V1 from 5.18 m to 0.24 m. Both Fig.\ref{fig9}(b) and TABLE I indicate that stop line information can greatly help the self-positioning for the first stopped vehicle. 

As previously mentioned, compared with the self-positioning method, CP sightly increased the errors of V1 without the stop line information. But the SL-CP significantly reduced the positioning error of V1 from around 6 m to less than 0.5 m during ${t_{s1}} \sim {t_e}$, which is clearly shown in Fig.\ref{fig9}(b). As shown from TABLE I, the RMSE of V1 during this period decreased from 7.10 m to 0.25m, which achieved a significant improvement from the CP method. More importantly, the improvement of V1 can further help the other two vehicles with the proposed SL-CP framework. Fig.\ref{fig9}(a) and (c) show that during ${t_{s1}}\sim t_e $, SL-CP dramatically improved the positioning performance of both V0 and V2. Their RMSEs (see TABLE I) reduced from 5.10 m and 6.02 m to 1.93 m and 3.84 m, respectively.

The experimental results of this scenario indicate that the stop line can be used for the positioning enhancement of the first stopped vehicle. With the proposed SL-CP framework, the enhancement can be extended to the whole VANET.

\begin{figure*}[t!]  %全栏显示
\centering
\subfigure[the $t_0$ instant]{               
\includegraphics[scale=0.27]{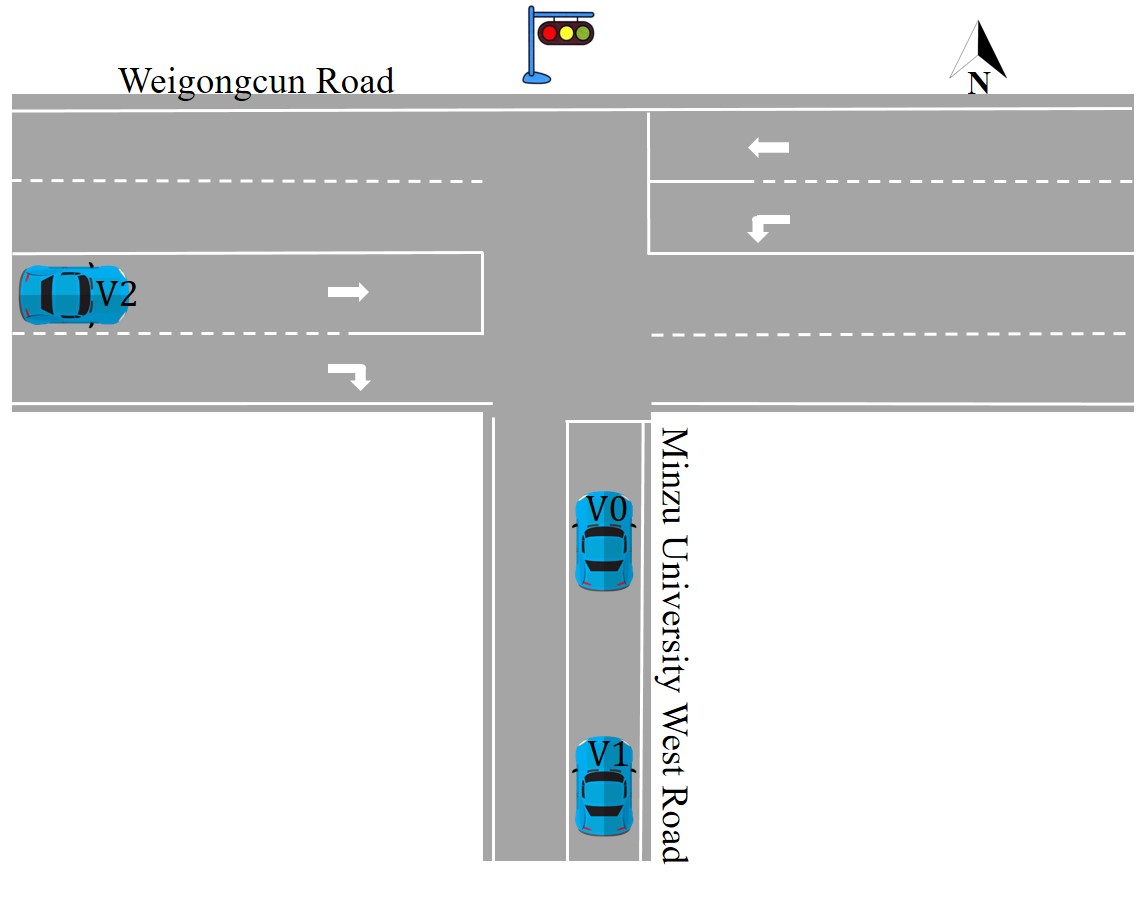}}
\hspace{0in}
\subfigure[the ${t_{s1}}$ instant]{
\includegraphics[scale=0.27]{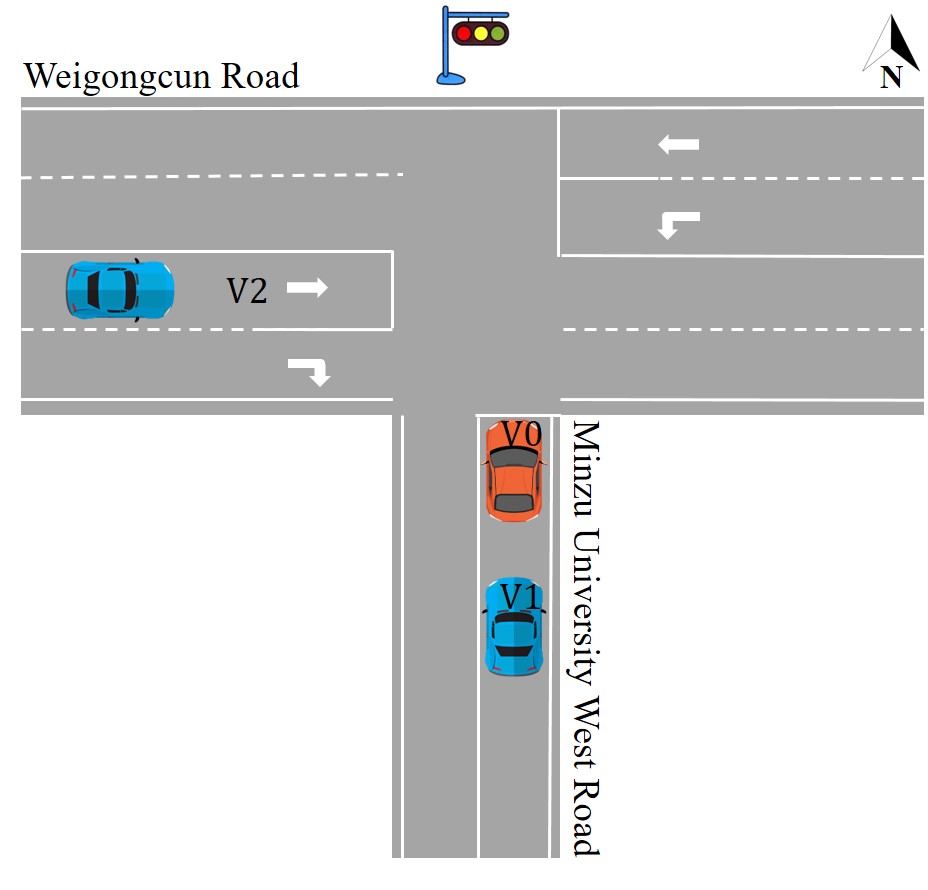}}
\hspace{0in}
\subfigure[the ${t_{s2}}$ instant]{
\includegraphics[scale=0.27]{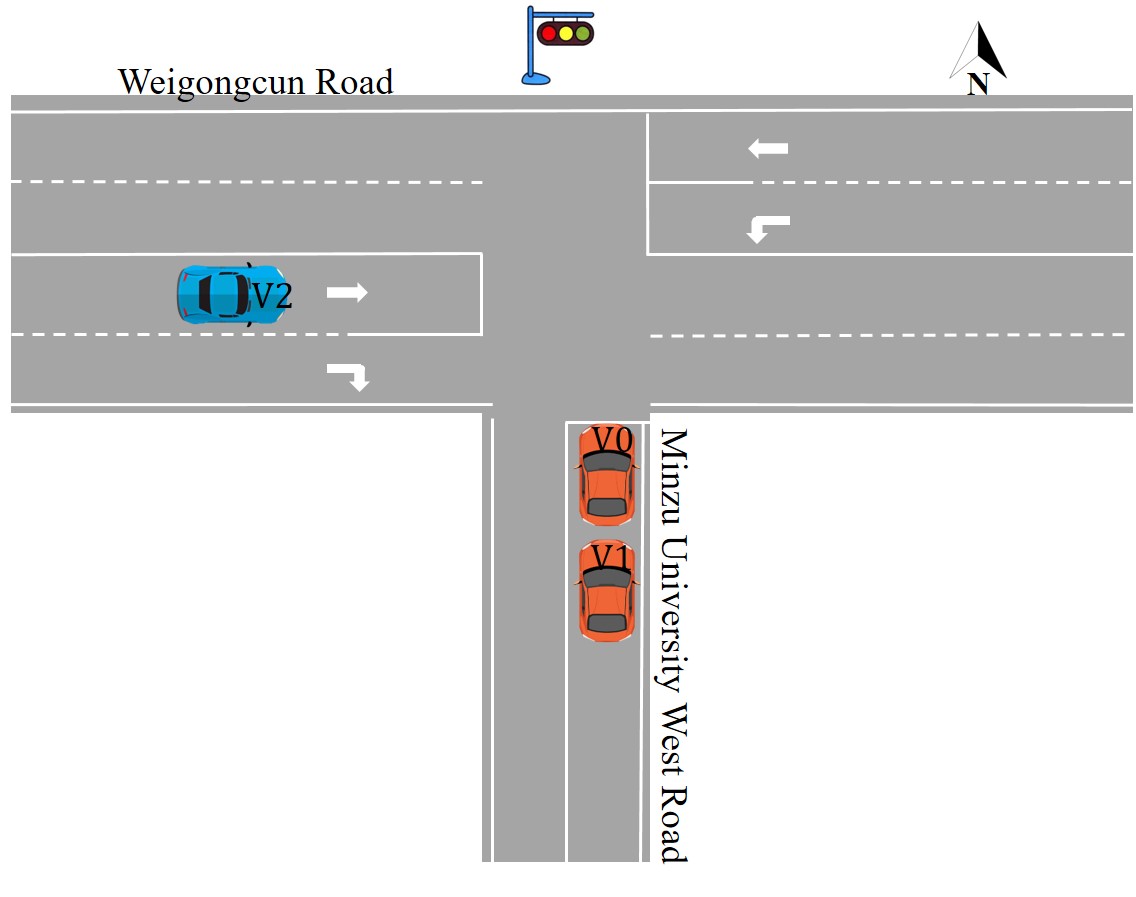}}

\subfigure[the ${t_{s3}}$ instant]{               
\includegraphics[scale=0.23]{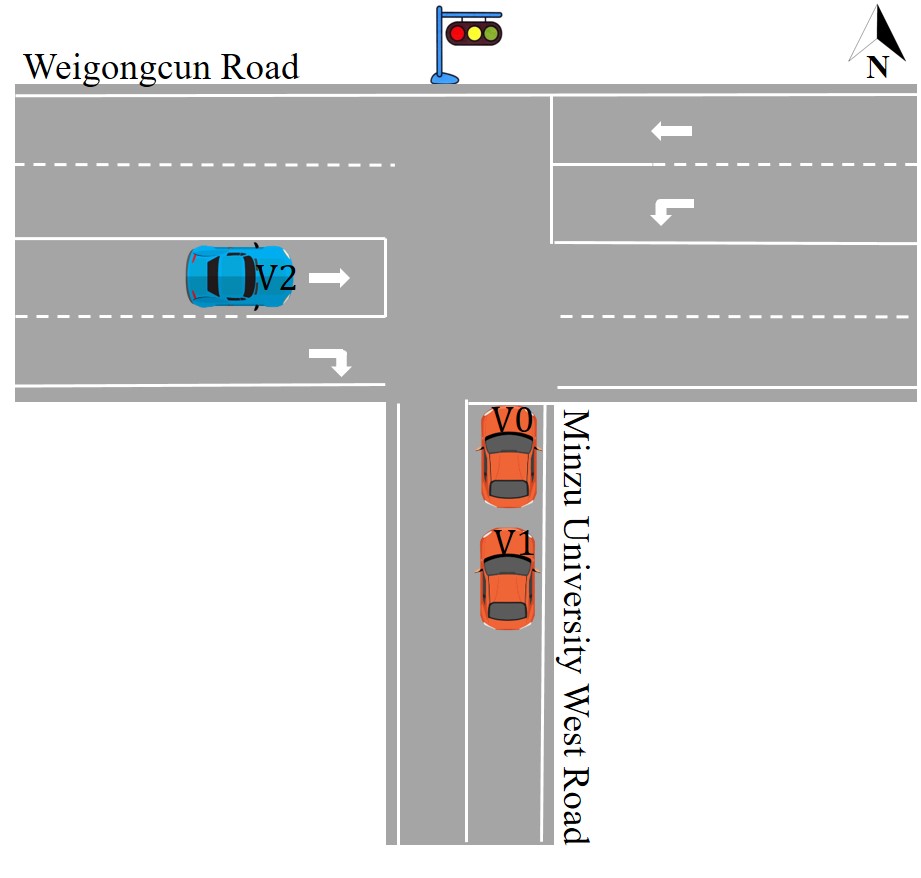}}
\hspace{0in}
\subfigure[the ${t_{s4}}$ instant]{               
\includegraphics[scale=0.23]{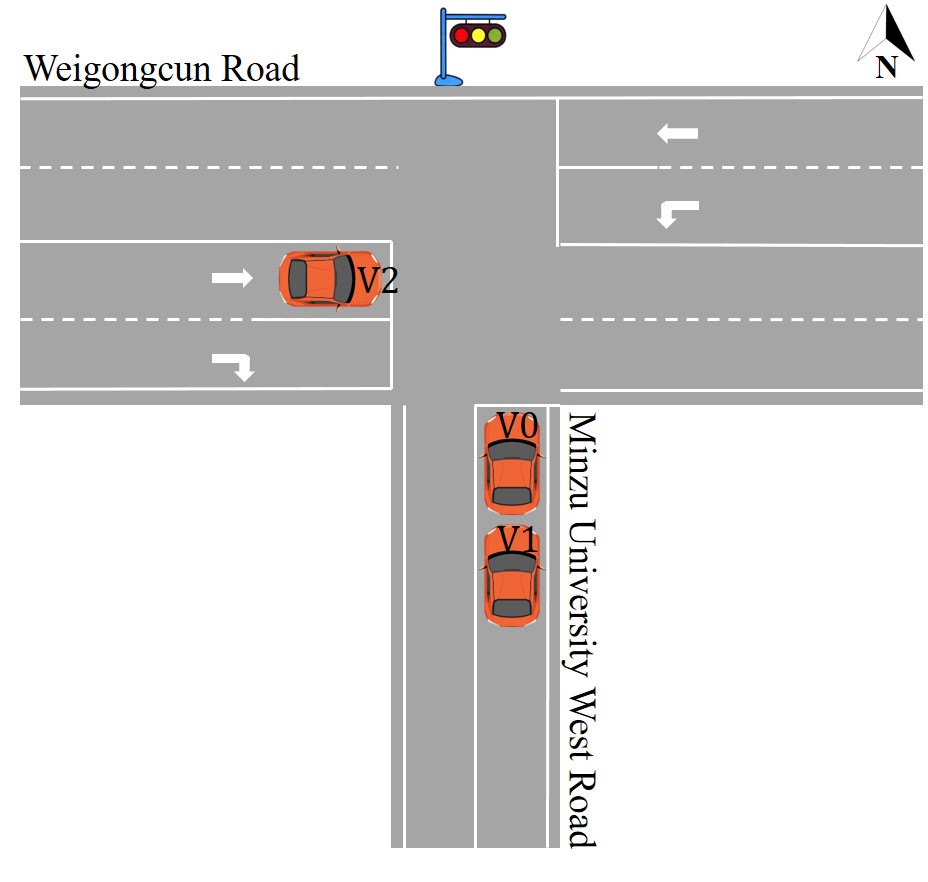}}
\hspace{0in}
\subfigure[the ${t_{s5}}$ instant]{
\includegraphics[scale=0.23]{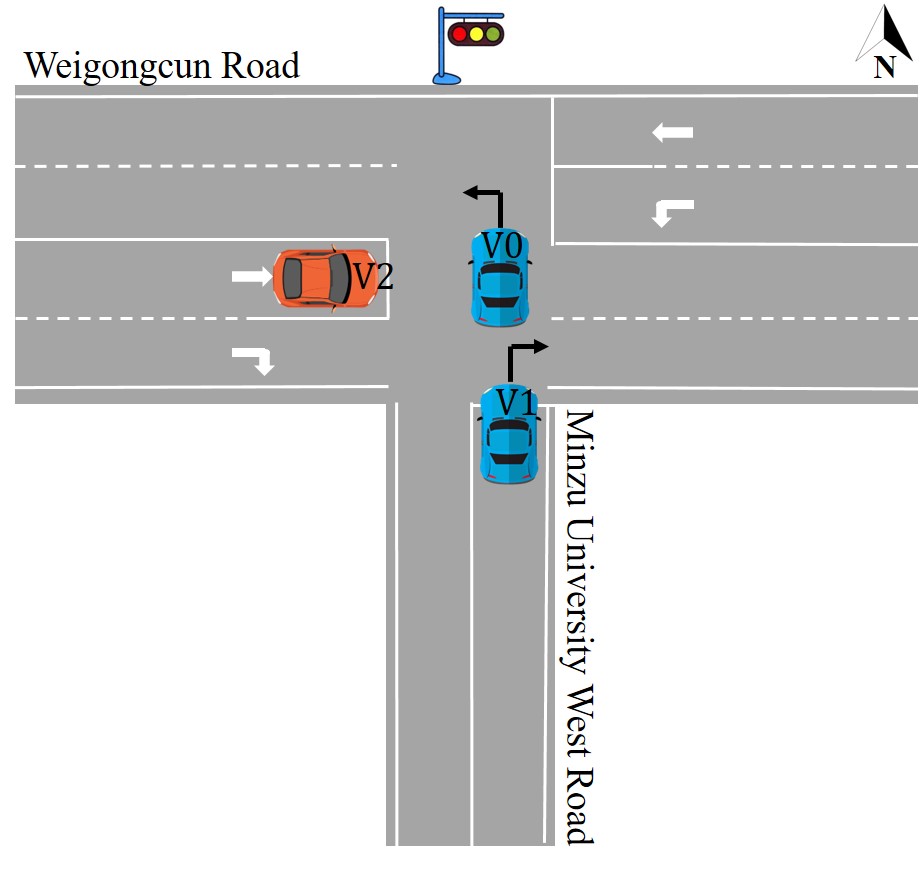}}
\hspace{0in}
\subfigure[the ${t_e}$ instant]{
\includegraphics[scale=0.23]{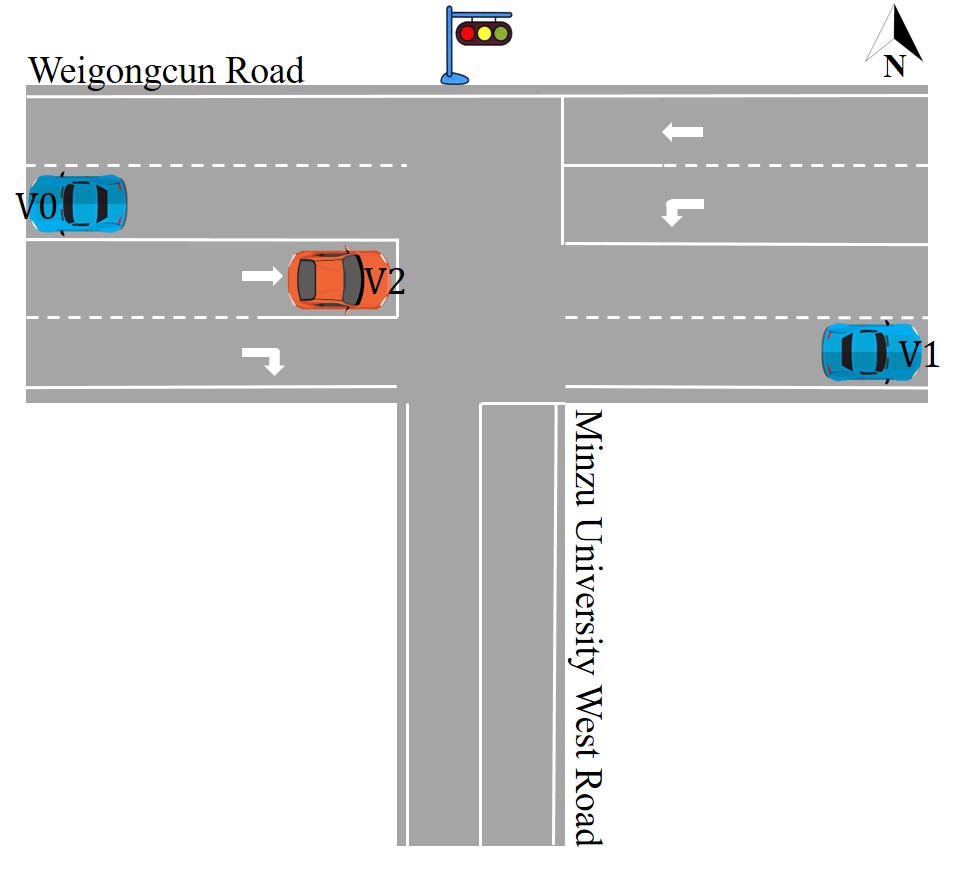}}
\caption{The key scenes of the scenario 2. Orange represents the vehicle is stopped. Blue represents the vehicle is moving.}
\label{fig10}
\end{figure*}

\begin{figure*}[]  %全栏显示
\centering
\subfigure[V0]{               
\includegraphics[scale=0.25]{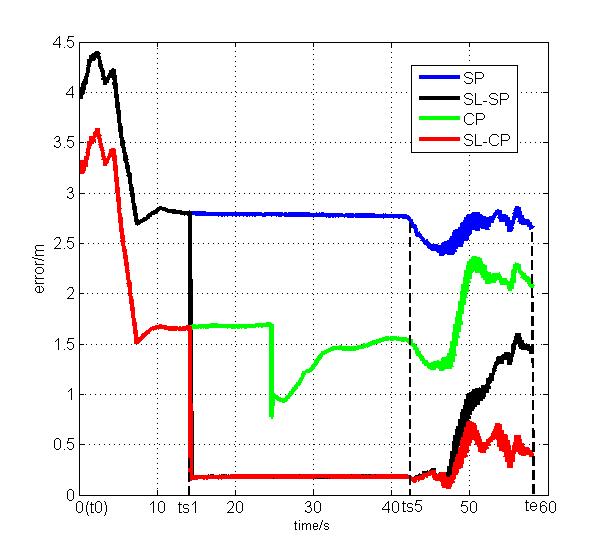}}
\hspace{0in}
\subfigure[V1]{
\includegraphics[scale=0.25]{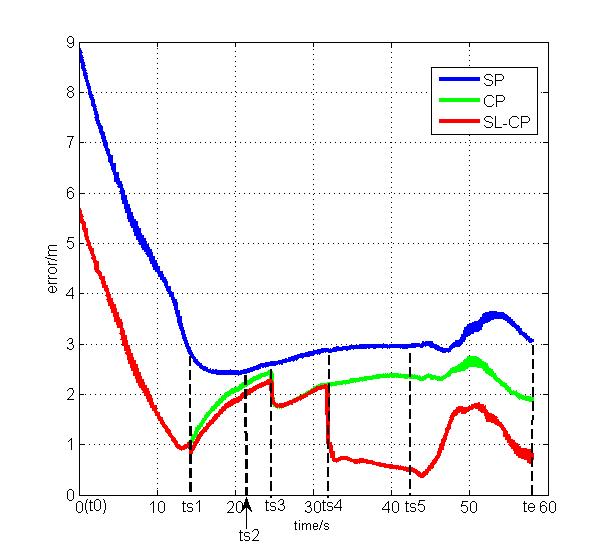}}
\hspace{0in}
\subfigure[V2]{
\includegraphics[scale=0.25]{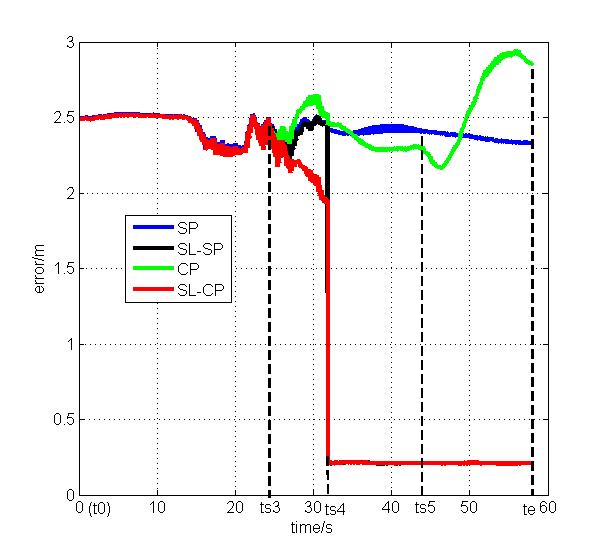}}
\caption{The positioning errors of the three vehicles in scenario 2.}
\label{fig11}
\end{figure*}

\subsection{Experimental Results for Scenario 2}	
		
We then implement the other scenario to further validate the proposed framework. When traffic lights switch, the role of the first stopped vehicle also shifts from one vehicle to another. In this scenario, two vehicles travel from south to north on Minzu University West Road, and the third vehicle goes straight from west to east on Weigongcun Road.

Fig.\ref{fig10} provides the critical scenes of this scenario. Fig.\ref{fig10}(a) shows V0 and V1 were running on the Minzu University West Road, and V2 was moving on the Weigongcun Road at the ${t_0}$ instant. Fig.\ref{fig10}(b) shows that V0 became the first stopped vehicle while V1 and V2 were still running at the ${t_{s1}}$ moment. Fig.\ref{fig10}(c) illustrates that V1 stopped to wait for the green light.
At ${t_{s3}}$, V2 ran into the line-of-sight (LOS) range of V0 and V1. All three vehicles could measure relative distances and communicate with each other. During ${t_0} \sim {t_{s3}}$, V2 could not stably communicate with either V0 or V1. At ${t_{s4}}$, V2 became the first stopped vehicle. As shown in Fig. \ref{fig10}(f), at ${t_{s5}}$, the traffic light turned green, V0 was turning left, and V1 was turning right. Fig.\ref{fig10}(g) shows that at ${t_{e}}$, V2 was still stopped, and both V0 and V1 were far away. After ${t_e}$, the communications between the three vehicles might not be stable due to the limitation of communication distance.

\begin{table}[htbp]
\centering
\caption{POSITIONING ERROR (RMSE) of V0 and V1 during ${t_{s1}} \sim {t_{s5}}$}
\begin{tabular}{p{1.2cm}p{1.2cm}p{1.2cm}p{1.2cm}p{1.2cm}}%l=left, r=right,c=center分别代表左对齐，右对齐和居中，字母的个数代表列数
\hline
 &SP &SL-SP &CP &SL-CP\\ \hline
V0 &2.78 m &0.27 m &1.49 m &0.26 m\\ 
V1 &2.73 m  &/ m &2.09 m &1.52 m\\\hline
\end{tabular}
\end{table}

\begin{table}[htbp]
\centering
\caption{POSITIONING ERROR (RMSE) of V0 and V1 during ${t_{s5}} \sim {t_{e}}$}
\begin{tabular}{p{1.2cm}p{1.2cm}p{1.2cm}p{1.2cm}p{1.2cm}}%l=left, r=right,c=center分别代表左对齐，右对齐和居中，字母的个数代表列数
\hline
 &SP &SL-SP &CP &SL-CP\\ \hline
V0 &2.64 m &0.95 m &1.88 m &0.41 m\\ 
V1 &3.20 m  &/ m &2.34 m &1.23 m\\\hline
\end{tabular}
\end{table}

\begin{table}[htbp]
\centering
\caption{POSITIONING ERROR (RMSE) of V2}
\begin{tabular}{p{1.2cm}p{1.2cm}p{1.2cm}p{1.2cm}p{1.2cm}}%l=left, r=right,c=center分别代表左对齐，右对齐和居中，字母的个数代表列数
\hline
 &SP &SL-SP &CP &SL-CP\\ \hline
${t_{s1}} \sim {t_{s4}}$ &2.38 m &2.38 m &2.41 m &2.28 m\\ 
${t_{s4}} \sim {t_{e}}$ &2.39 m  &0.26 m &2.49 m &0.28 m\\\hline
\end{tabular}
\end{table}

The positioning errors (RMSE) are shown in Fig.\ref{fig11}, and the detailed results are shown in TABLE II $\sim$ IV. During ${t_0} \sim {t_{s1}}$, no vehicle was stopped, and only V0 and V1 could mutually communicate and measure the relative distance. Hence, the positioning results of V0 and V1 calculated by the SL-CP and the CP methods were identical (see Fig.\ref{fig11}). Compared with self-positioning, using SL-CP or CP decreased the positioning errors of both V0 and V1 significantly during that period.

At ${t_{s1}}$, V0 became the first stopped vehicle. With SL-SP, the localization error directly declined to less than 0.5 m (red curve in Fig.\ref{fig11}(a)). During ${t_{s1}} \sim {t_{s5}}$, V0 waited for the green light, and the RMSE of the SL-SP was 0.27 m (see TABEL II), which could provide much better initialization for the localization of V0 after ${t_{s5}}$. At ${t_{s5}}$, V0 started to move. The RMSE of SL-SP for the time slot ${t_{s5}}\sim t_e$ was 0.95 m, much smaller than that of the common self-positioning due to the better initialization. Compared to SL-SP,  SL-CP can further benefit from cooperation with the other two vehicles, especially the new first stopped vehicle, i.e., V2. The RMSE reduced to 0.41m (see TABEL III). With the help of the stop line and the SL-CP scheme, V0 significantly improved the overall positioning performance.

At ${t_{s3}}$, V2 ran into the LOS range of V0 and V1. During ${t_{s3}} \sim {t_{s4}}$, the CP of V2 was slightly worse than the self-positioning while SL-CP had a dramatic improvement due to the better self-positioning of V0, i.e., the first stopped vehicle.
At ${t_{s4}}$, V2 became the first stopped vehicle and the positioning error was greatly reduced with the help of stop line. As can be seen from TABEL IV, the RMSE of SL-SP is 0.26 m and that of SL-CP was 0.28 m during ${t_{s4}} \sim {t_{e}}$. In the same period, ignoring the stop line information, the positioning errors of SP and CP were both more than 2 m.

Within the SL-CP framework, V1 can greatly benefit from the other two vehicles. At ${t_{s1}}$, V0 became the first stopped vehicle, and the positioning error of V1 via the SL-CP had a modest reduction. At ${t_{s4}}$, V2 also became the first stopped vehicle, and consequently V1's positioning error was further reduced. In the time slot ${t_{s1}} \sim {t_{s5}}$, compared with the CP method, the RMSE of V1 via SL-CP decreased from 2.09 to 1.52 m. At ${t_{s5}}$, V0 and V1 began to move one after another, and hence V0 could not adopt the stop line information for self-positioning enhancement. Therefore, the improvement of V1 introduced by V2V communication experienced a decreasing trend. Nonetheless, compared with the CP method, the RMSE of V1 via the SL-CP reduced from 2.34 m to 1.23 m during ${t_{s5}} \sim {t_e}$ due to the new stopped vehicle, i.e., V2.

In this scenario, vehicles successively stopped due to the red light. The role of the first stopped vehicle shifted from one vehicle to another when the traffic lights switched. It is a more general scenario in heavy-traffic areas. The experiment results show the effectiveness of the stop-line-aided self-positioning and the enhanced performance of the proposed SL-CP framework for the whole VANET in such a general intersection scenario.

\begin{figure}[]  
\centering
\subfigure[Scenario 1-V0]{               
\includegraphics[scale=0.2]{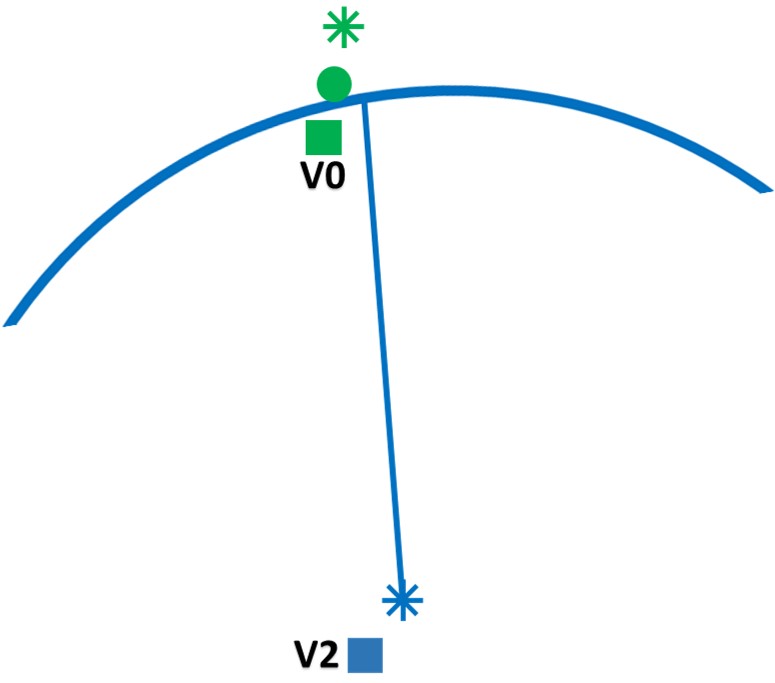}}
\hspace{0.5in}
\subfigure[Scenario 1-V2]{
\includegraphics[scale=0.2]{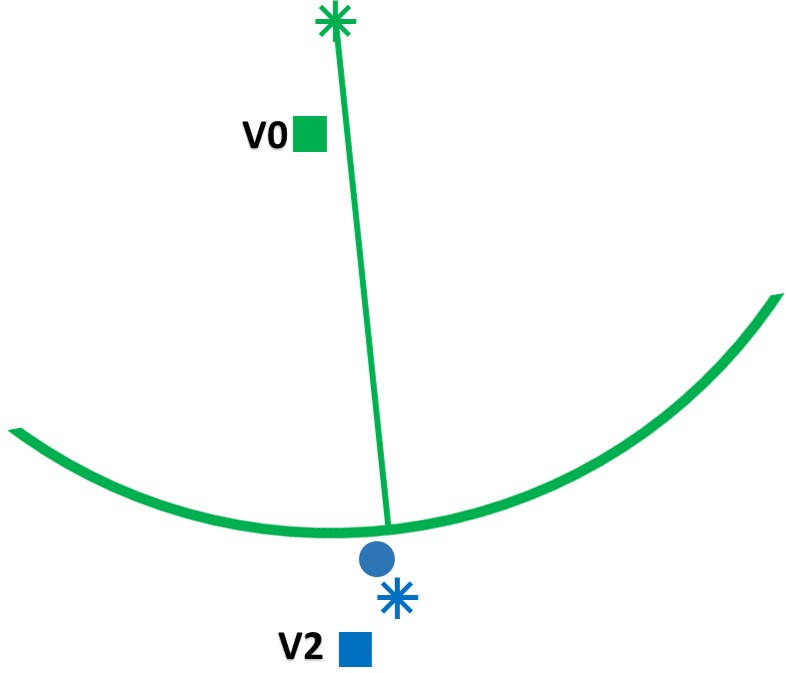}}

\subfigure[Scenario 2-V0]{               
\includegraphics[scale=0.2]{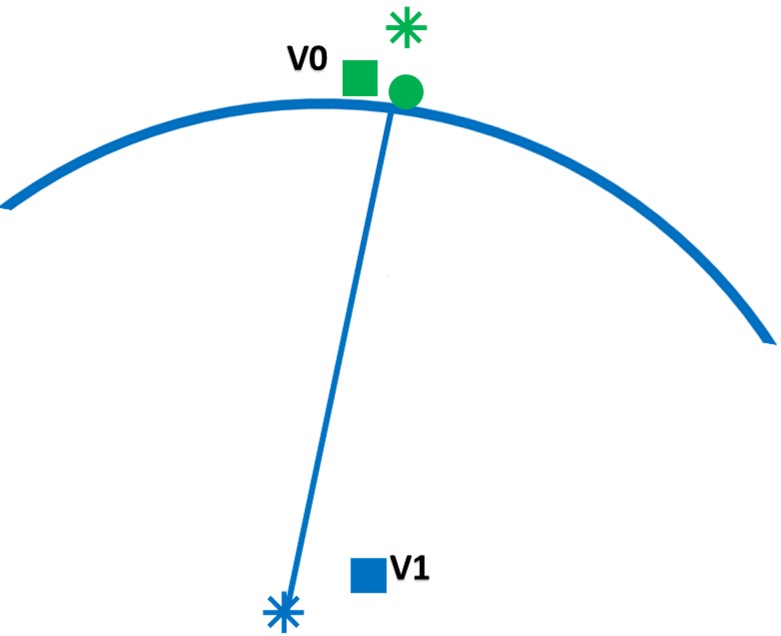}}
\hspace{0.5in}
\subfigure[Scenario 2-V1]{
\includegraphics[scale=0.2]{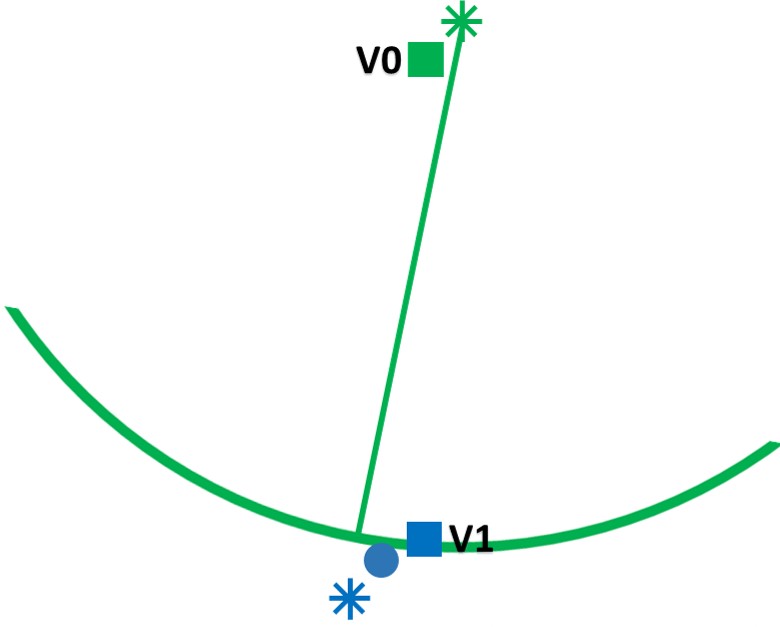}}

\caption{Illustration for cooperative positioning: square represents the real position, star represents the self-positioning result, and the circle represents the cooperative positioning result.}
\label{error}
\end{figure}

\subsection{Discussion}

In scenario 2, compared with the self-positioning method, the positioning errors of both V0 and V1 were decreased during ${t_{0}} \sim {t_{s3}}$. 
In scenario 1, however, CP reduced the positioning error of V0 while increasing the errors V2 during ${t_{0}} \sim {t_{e}}$. 
As shown in Fig.\ref{error}(a),  CP method fuses the self-positioning results with the V2V ranging, i.e., ${\hat z_{2 \to 0}}$, then the estimated positioning result of V0 is closer to the real position. But Fig.\ref{error}(b) shows the opposite case for V2. Fig.\ref{error}(c) and (d) illustrate that the CP results of V0 and V1 are both closer to the real position over the self-positioning method in scenario 2. Fig. \ref{error} indicates that the error direction of the self-positioning result is related to the performance of cooperative positioning.  

It has the potential to further improve the performance by ignoring certain vehicles in cooperative positioning algorithm. For example, in scenario 1, it is better for V2 to ignore V0 in both cooperative positioning methods. However, this phenomenon is out of the scope of this work and will be discussed in our further work.

\section{Concluding remarks and future works}

This paper develops a novel stop-line-aided cooperative inertial navigation framework to enhance the localization of connected vehicles in intersection scenarios. The stop line information is introduced to improve the first stopped vehicle, and the CIN framework is applied to further extend the improvement to the whole VANET. The experimental results indicate \begin{enumerate}
    \item The stop line information can significantly enhance the self-positioning of the first stopped vehicle; 
    \item The proposed stop line aided cooperative positioning (SL-CP) framework can enhance the positioning of the whole VANET in intersection scenarios. 
\end{enumerate}

In the further, we will focus on the node-selection mechanism in the cooperative positioning framework to detect and remove the low informative nodes, for example, V0 in scenario 1, to further improve the overall performance of a VANET.

\ifCLASSOPTIONcaptionsoff
  \newpage
\fi

\end{document}